\documentclass[lettersize,journal]{IEEEtran}
\usepackage{amsmath,amsfonts,amssymb,bm}
\usepackage{array}
\usepackage[caption=false,font=normalsize,labelfont=sf,textfont=sf]{subfig}
\usepackage{stfloats}
\usepackage{verbatim}
\usepackage{algorithm}
\usepackage{textcomp}
\usepackage{xcolor}
\usepackage{graphicx}
\usepackage{amssymb}
\usepackage{algpseudocode}
\usepackage{float}
\usepackage{cite}
\usepackage{url}
\usepackage{booktabs}
\usepackage{multirow}
\usepackage{hyperref}
\usepackage{amsmath}
\usepackage{physics}
\renewcommand{\arraystretch}{1.2}

\def\BibTeX{{\rm B\kern-.05em{\sc i\kern-.025em b}\kern-.08em
    T\kern-.1667em\lower.7ex\hbox{E}\kern-.125emX}}
\usepackage{balance}
\begin{document}
\title{Towards On-Device Learning and Reconfigurable Hardware Implementation for Encoded Single-Photon Signal Processing}

\author{Zhenya Zang,
        Xingda Li,
        and David Day Uei Li
\thanks{This work is supported by the EPSRC (EP/T00097X/1); the Quantum Technology Hub in Quantum Imaging (QuantiC), and the University of Strathclyde. Xingda Li also acknowledges support from China Scholarship Council.}
\thanks{Zhenya Zang, Xingda Li, and David Day Uei Li are with the Department of Biomedical Engineering, University of Strathclyde, Glasgow, UK. (e-mail: zhenya.zang@strath.ac.uk, xingda.li@strath.ac.uk, david.li@strath.ac.uk).}
}

\markboth{Manuscript submitted using IEEE \LaTeX\ template}%
{Your Name \MakeLowercase{\textit{et al.}}: Towards On-Device Learning and Reconfigurable Hardware Implementation}

\maketitle

\begin{abstract}
Deep neural networks (DNNs) enhance the accuracy and efficiency of reconstructing key parameters from time-resolved photon arrival signals recorded by single-photon detectors. However, the performance of conventional backpropagation-based DNNs is highly dependent on various parameters of the optical setup and biological samples under examination, necessitating frequent network retraining—either through transfer learning or from scratch.  Newly collected data must also be stored and transferred to a high-performance GPU server for retraining, introducing latency and storage overhead. To address these challenges, we propose an online training algorithm based on a One-Sided Jacobi rotation-based Online Sequential Extreme Learning Machine (OSOS-ELM). We fully exploit parallelism in executing OSOS-ELM on a heterogeneous FPGA with integrated ARM cores. Extensive evaluations of OSOS-ELM and OS-ELM demonstrate that both achieve comparable accuracy across different network dimensions (i.e., input, hidden, and output layers), while OSOS-ELM proves to be more hardware-efficient. By leveraging the parallelism of OSOS-ELM, we implement a holistic computing prototype on a Xilinx ZCU104 FPGA, which integrates a multi-core CPU and programmable logic fabric. We validate our approach through three case studies involving single-photon signal analysis: sensing through fog using commercial single-photon LiDAR, fluorescence lifetime estimation in fluorescence lifetime microscopy, and blood flow index reconstruction in diffuse correlation spectroscopy—all utilizing one-dimensional data encoded from photonic signals. From a hardware perspective, we optimize the OSOS-ELM workload by employing multi-tasked processing on ARM CPU cores and pipelined execution on the FPGA’s logic fabric. We also implement our OSOS-ELM on the NVIDIA Jetson Xavier NX GPU to comprehensively investigate its computing performance on another type of heterogeneous computing platform.
\end{abstract}

\begin{IEEEkeywords}
Online learning neural networks, reconfigurable hardware, time-resolved single-photon signal processing
\end{IEEEkeywords}

\section{Introduction}
\IEEEPARstart{O}{n}-device training of neural networks has been emerging in recent decades. On-device training and inference save the overhead of data transfer to data centers, memory management, and computing on the cloud. The number of edge devices is increasing exponentially and is expected to reach 1 trillion by 2035~\cite{Sparks2021}. Latency tends to be a bottleneck of real-time applications such as healthcare and machine automation. Additionally, information privacy can be threatened when uploading and offloading sensitive biomedical data to the cloud. Implementing robust deep neural networks (DNNs) for local training and inference is crucial to addressing these issues. Industry~\cite{lai2018cmsis} and academia~\cite{lin2022device} proposed efficient network compression techniques and kernel optimization of DNNs on ARM-based processors to realize real-time training. Alongside backpropagation-based DNNs, extreme learning machines (ELMs) are compact, backpropagation-free, single-layer forward networks (SLFNs), demonstrating high accuracy and compact model size for wearable biometric signal processing (EEG~\cite{khare2022vhers}, ECG~\cite{chen2023s}, and PPG~\cite{mahuli2016prediction}) and industrial signal processing (energy supply~\cite{shi2021novel}, power quality disturbances~\cite{zhao2018novel}). Additionally, ELM is robust in photonic sensor-based applications, including sensing through fog~\cite{zang2024object}, fluorescence lifetime imaging (FLIM)~\cite{zang2022fast}, and diffuse correlation spectroscopy (DCS)~\cite{zang2025fast} for encoded single-photon signal processing. Moreover, ELM variants like Multilayer-ELM~\cite{tang2015extreme} and Online Sequential (OS)-ELM~\cite{huang2005line} provide a baseline for online learning, allowing incremental learning of new samples with labels without forgetting mechanisms. In contrast, conventional DNNs require augmenting new samples into old datasets and retraining, either from scratch or through transfer learning.

\begin{figure*}[t]
    \centering
    \includegraphics[width=\linewidth]{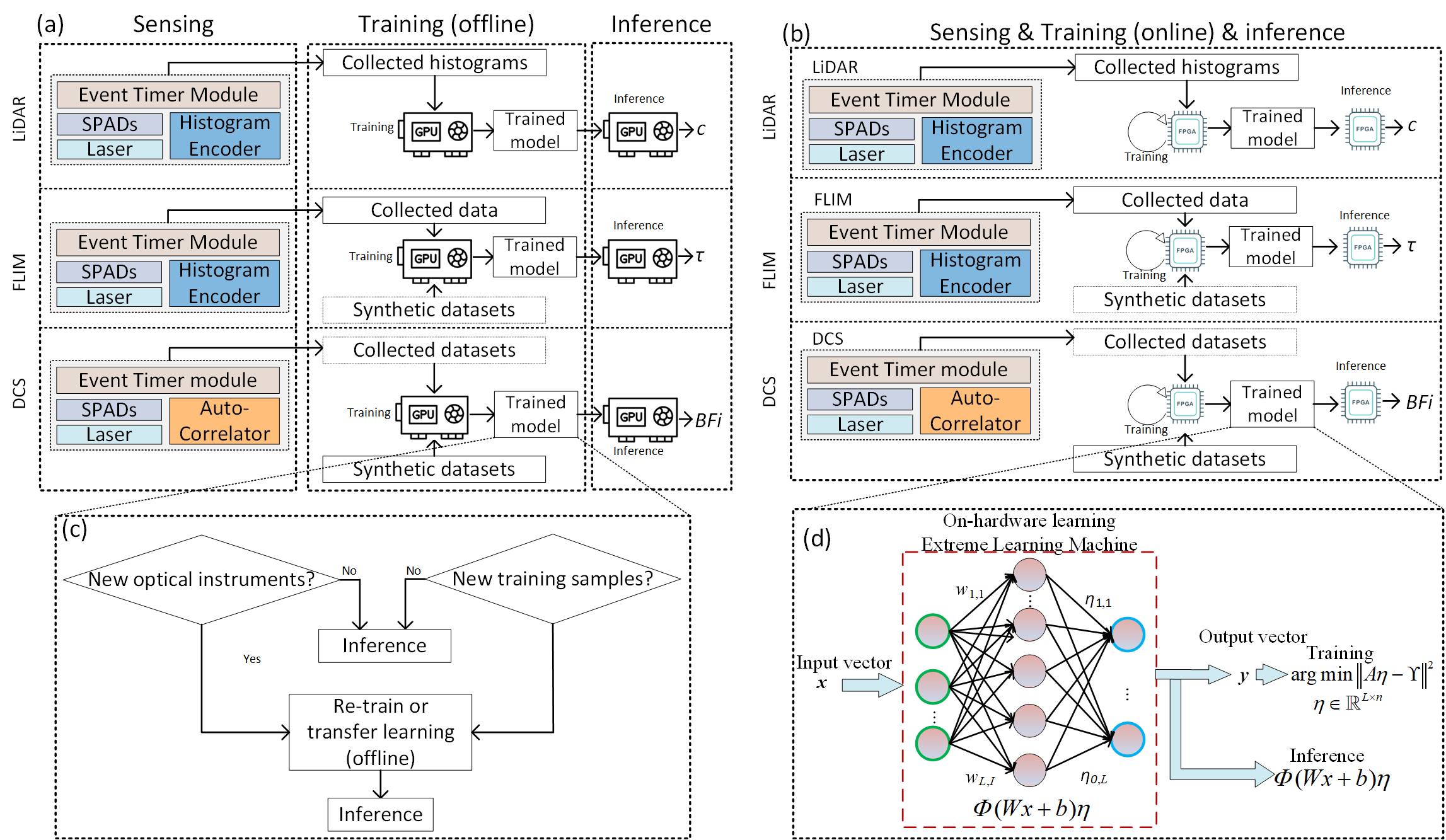}
    \caption{Dataflow of data acquisition and processing for FLIM, DCS, and LiDAR: (a) Optical instruments, where histogram encoding and auto-correlator modules are integrated into the optical systems for single-photon LiDAR, FLIM, and DCS, along with standard single-photon detectors (using SPADs as an example) and timer modules. Additionally, offline training pipelines for FLIM and DCS are included. \( \tau \), \( c \), and BFi indicate object class, fluorescence lifetime, and blood flow index. (b) FPGA-based ELM online training. (c) A flowchart illustrating the process for handling newly acquired data from a new experimental system or with new optical parameters. (d) OSOS-ELM training and inference dataflow (the topologies between OSOS-ELM and ELM are the same).}
    \label{fig:figure1_overview}
\end{figure*}

ELM demonstrates superior accuracy and speed compared to iterative optimization algorithms. It also outperforms DNNs by reducing training dependency on various experimental factors, avoiding local minima during training, and alleviating overfitting. This advantage is particularly evident in FLIM and DCS, as they require a fast response to incoming data after being trained on large datasets for a specific experimental platform. FLIM and DCS share similar optical setups and rely on powerful post-processing hardware for parameter reconstruction. Similarly, direct time-of-flight (dToF) LiDAR uses similar components but is used for depth reconstruction. This work adopts the cases to demonstrate our novel improved ELM algorithm and its corresponding hardware computing architecture for training and inference. Comparisons of the experimental setups among typical FLIM, DCS, and LiDAR, followed by conventional DNN- and ELM-based pre-processing for parameter reconstruction, are depicted in Fig.~\ref{fig:figure1_overview}. dToF single-photon LiDAR generates ToF values from encoded histograms. Therefore, besides regression tasks for FLIM and DCS, we also employ classification for single-photon LiDAR to enable object recognition in foggy conditions. Fig.~\ref{fig:figure1_overview} describes data acquisition and DNN-based processing pipelines. Conversely, the concept we propose in Fig.~\ref{fig:figure1_overview}(b) integrates the three stages (data acquisition, training, and inference), saving time and computing overhead of retraining. In Fig.~\ref{fig:figure1_overview}(c), although transfer learning~\cite{torrey2010transfer} can leverage pre-trained models and update existing training parameters, backpropagation on a GPU with sufficient on-chip memory and bandwidth is still necessary for offline training. DNNs struggle to learn new samples alone, degrading accuracy due to learning bias towards new samples. Therefore, this study aims to bypass backpropagation and optimize existing OS-ELM by using multi-threading and operation parallelism to accelerate training on FPGA. We integrated one-sided Jacobi rotation singular value decomposition (OJR-SVD)~\cite{takata2021implementation} into the pseudo-training (big matrix pseudo-inversion) strategy in OS-ELM, termed OSOS-ELM, making it feasible for multi-threading on hardware for local training. Fig.~\ref{fig:figure1_overview}(d) illustrates the training and inference dataflow of OSOS-ELM. The FLIM and DCS training datasets are generated based on rigorous analytical models from our previous studies~\cite{zang2022fast, zang2025fast}. The training process is implemented using our proposed OJR-SVD method and MATLAB's built-in SVD to compare performance. Besides, the LiDAR datasets are collected from real-world scenarios.

Models generating synthetic fluorescence decays for FLIM~\cite{becker2012fluorescence} and autocorrelation functions (ACFs) of DCS~\cite{durduran2014diffuse} follow the sensing principle using pulsed and continuous-wave lasers and single-photon detectors. Single-photon avalanche diodes (SPADs) are emerging in LiDAR~\cite{chen2020data}, FLIM~\cite{xiao2021dynamic}, and DCS~\cite{della2023field,della2020128} due to their high single-photon sensitivity. When integrated with SPAD-based systems, FPGAs become critical electronic components to configure working modes~\cite{henderson2019192}, provide clocks for refreshing frames and lines of pixels~\cite{gyongy2020high}, measure precise time-of-flight for single photons~\cite{portaluppi2022multi}, and encode histograms on-chip~\cite{della2020128}. Also, FPGAs are increasingly being used as processing units~\cite{li2009fpga, della2023field, wayne2023massively} for SPADs due to the growing number of on-chip hardware resources.

DNNs are proven high-accuracy inference for lifetime ($\tau$) and blood flow index (BFi) reconstruction and are more robust to low signal-to-noise ratios (SNRs)~\cite{zang2023compact, smith2019fast, chen2022generative, chen2023spatial, wang2024quantification}. Therefore, DNN inference on FPGA has been applied to SPAD-based sensing systems~\cite{afshar2020event, lin2024coupling, zang2023compact, xiao2021dynamic}. However, DNN implementations with learned parameters on FPGAs are tied to specific electronic and optical setups, making the models infeasible to transfer to other setups, let alone different applications. Additionally, existing work is only capable of forward propagation rather than training.

In this work, we implement OSOS-ELM on FPGA to deliver a platform- and application-independent hardware solution for training and inference for encoded single-photon signals. From existing on-chip histogramming~\cite{mai2023development} and ACF~\cite{della2023field} computing techniques of FLIM and DCS, on-chip training and inference can be interfaced with such on-chip preprocessing methods.

The contribution of this work is threefold:
\begin{enumerate}
    \item We propose OSOS-ELM to resolve the computationally intensive Moore--Penrose inverse of big matrices during training. The training procedures are decomposed into independent tasks, presenting a high potential for multi-tasking and parallelism. We demonstrate that our OSOS-ELM can perform holistic training, including initial training (IT) and one-batch training (OBT), on a system-on-chip (SoC)-based FPGA.
    \item We evaluate the accuracy and speed of OSOS-ELM on FLIM, DCS, and LiDAR datasets for training and inference. The results demonstrate that OSOS-ELM exhibits no performance degradation compared to conventional software implementations of SVD-based OS-ELM in MATLAB while achieving higher parallelism. We analyze different factors—the number of initial batches ($N_0$), the batch size of sequential training, and the number of hidden nodes ($L$)—to quantify the reconstruction error and latency. Additionally, we quantitatively compare the accuracy of OSOS-ELM implemented with floating-point (FLP) and fixed-point (FXP) arithmetic on FLIM and DCS datasets for training and inference, providing a reference design for hardware.
    \item We implement the training and inference of OSOS-ELM on a ZCU104 Ultrascale+ MPSoC FPGA by leveraging interruption-enabled, multi-task processing on multi-CPU cores on the processing system (PS) and parallelized processing on programmable logic (PL). The topology of the network, the number of nodes in the input layer (\#IN), hidden layer ($L$), output layer (\#ON), and the parameters of converging conditions are parameterized from the PS. Working modes (training or inference) and parameter preloading instructions are also reconfigurable from the PS. The latency and power consumption of our hardware implementation are compared with those of a GPU.
\end{enumerate}

\section{Prior Work}
This section reviews existing work on ELM online training using FPGA-based processors for processing encoded single-photon signals from single-photon detectors.

\subsection{ELM Implementation on Hardware}
As the pseudo-inversion of matrices is the essential training strategy for naive ELM, researchers focus on efficient hardware implementation for it. Frances-Villora \textit{et al.}~\cite{frances2018moving} presented the first OS-ELM processor implemented on FPGA, which is compatible with various numbers of nodes ($L$) in the hidden layer with high learning frequency (14~kHz for $L = 40$ to 180~Hz for $L = 500$). Safaei \textit{et al.}~\cite{safaei2018system} implemented Gram--Schmidt-based QR-decomposition (QRD) on FPGA with deep pipelining and efficient memory allocation, supporting both ELM and OS-ELM. Tsukada \textit{et al.}~\cite{tsukada2020neural} implemented an OS-ELM with a lightweight forgetting mechanism on a PYNQ-Z2 SoC using high-level synthesis (HLS) to detect anomalies from a normal distribution with high computation efficiency.

A spiking ELM~\cite{he2021low} variant was equipped with on-chip triplet-based reward-modulated spike-timing-dependent plasticity to perform training and inference with low hardware consumption using encoded spikes among layers. Decherchi \textit{et al.}~\cite{decherchi2012efficient} explored the sparsity of the loss function of ELM, thereby pruning neurons and saving hardware consumption. The pruned ELM was implemented on an FPGA to perform as a classifier. Sahani \textit{et al.} used Simulink to implement Hilbert--Huang transform-based ELM~\cite{sahani2019fpga} and P-Norm-ELM~\cite{sahani2019automatic} on FPGA to monitor power quality disturbance and recognize power quality events. Chen \textit{et al.}~\cite{chen2023s} designed an application-specific integrated circuit (ASIC) of ELM for ECG signal classification, using scalable QRD to achieve online incremental learning. Huang \textit{et al.}~\cite{huang2021generic} reported a parallelized and memory-optimized recursive least mean p-power extreme learning machine (RLMP-ELM), exhibiting high accuracy and computational efficiency in classification tasks.

For ASIC implementation, Chuang \textit{et al.}~\cite{chuang2021arbitrarily} designed an ASIC implementing adaptive boosting and eigenspace denoising ELM for ECG anomaly detection. Chen \textit{et al.}~\cite{chen2015128} presented an analog-signal ASIC implementing ELM for a multi-channel brain-machine interface. Bataller-Mompean \textit{\textit{\textit{et al.}}}~\cite{bataller2016support} presented a software-hardware co-design framework that runs ELM for real-time detection of brain areas during electrode positioning in deep brain stimulation surgery. Li \textit{et al.}~\cite{li2019robust} proposed sub-eigenspace-ELM by leveraging compressed sensing to denoise input data, where an ASIC was taped out for ECG monitoring. A neuromorphic implementation of ELM~\cite{liang2019memristive} using memristors achieved the same classification accuracy as software. Another neuromorphic ELM processor, implemented with spintronic memristor-based synaptic circuits, biasing circuits, and activation function circuits, was proposed~\cite{dong2021neuromorphic}, which was successfully verified using a super-resolution image reconstruction.

\subsection{On-Chip Processing for FLIM, DCS, and LiDAR}
Zang \textit{et al.}~\cite{zang2023compact} implemented a compressed 1D convolutional neural network (CNN) on an FPGA to reconstruct amplitude- ($\tau_A$) and intensity-weighted ($\tau_I$) averaged lifetime for a time-correlated single-photon counting (TCSPC)-based FLIM system. A recurrent neural network was implemented on an FPGA and integrated into a TCSPC system to estimate fluorescence lifetime~\cite{lin2024coupling}. A compressed 1D-CNN was implemented on an FPGA for fast fluorescence lifetime sensing in a flow cytometry system~\cite{xiao2021dynamic}. A compact real-time FLIM system utilizing a SPAD array and an FPGA-implemented Integration for Extraction Method (IEM) demonstrates fast, background-insensitive fluorescence lifetime determination with a wide resolvability range, suitable for widefield FLIM~\cite{li2010real}. A center-of-mass algorithm was implemented in a SPAD array’s firmware for a flow cytometry system for imaging cell flow and identification with background compensation~\cite{mai2020flow}.

For FPGA-based computing modules for DCS, an autocorrelator was implemented in the firmware of a SPAD on an FPGA~\cite{della2023field} to achieve real-time ACF generation. An ACF fitting algorithm was implemented on an FPGA to process multi-channel avalanche photodetector in real-time~\cite{lin2019diffuse}. Two FPGAs were used to compute and compress ACFs for a SPAD array with high spatial resolution (500 $\times$ 500), achieving a 470$\times$ improvement in signal-to-noise ratio (SNR) compared to a single-detector system~\cite{wayne2023massively}. An autocorrelator array was implemented on an FPGA for a SPAD array-based fluorescence correlation spectroscopy enabling fast imaging~\cite{buchholz2012fpga}. Zang \textit{et al.}~\cite{zang2025towards} proposed a holistic hardware architecture implementing an ACF generator and a multiplication-free DNN for BFi reconstruction on the processing system (PS) and programmable logic (PL), respectively.

On-chip machine learning modules were implemented in an ASIC and integrated with SPAD-based LiDAR systems for super-resolution depth imaging~\cite{sun2022multi} and depth image reconstruction under low photon count conditions~\cite{cui202380}.

\section{Preliminaries}

This section introduces the fundamental concepts of ELM and OS-ELM, providing the foundations for OSOS-ELM. Additionally, it briefly introduces the data generation pipelines and analytical models of FLIM and DCS.

\subsection{ELM}
ELM is an SLFN that utilizes universal approximation theories to approximate versatile continuous functions given sufficient hidden neurons~\cite{huang2006extreme}. The weights of the input-to-hidden layer are randomly assigned and remain fixed during the training and inference.

Given $S$ training samples (i.e., $S$ pairs of training vectors $\bm{x} \in \mathbb{R}^{l \times m}$ and labels $\bm{y} \in \mathbb{R}^{l \times n}$, where $\bm{x}_i = [x_{i1}, x_{i2}, \dots, x_{im}]^T \in \mathbb{R}^m$ and $\bm{y}_i = [y_{i1}, y_{i2}, \dots, y_{in}] \in \mathbb{R}^n$ are the $i\textsuperscript{th}$ input and target vectors), suppose there are $L$ hidden nodes. The output matrix $\bm{A} \in \mathbb{R}^{l \times L}$ from the hidden layer is computed as $\bm{A} = \mathrm{\varPhi}(\bm{W} \bm{x} + \bm{b})$, where $\mathrm{\varPhi}$ is the activation function, typically sigmoid, for accurate training.

Here, $\bm{W} \in \mathbb{R}^{m \times L}$ is the randomized weight matrix between the input and hidden layers, and $\bm{b} \in \mathbb{R}^L$ is the bias vector in the hidden layer. The output of ELM is denoted by $\tilde{\bm{y}} = \mathrm{\varPhi}(\bm{W} \bm{x} + \bm{b}) \bm{\eta}$, where $\bm{\eta} \in \mathbb{R}^{L \times n}$ is the weight matrix to be trained between the hidden and output layers.

The Moore–Penrose pseudo-inverse is commonly used to learn $\bm{\eta}$, helping to avoid overfitting and the inefficiencies of iterative backpropagation. Accordingly, $\bm{\eta}$ is solved by $\hat{\bm{\eta}} = (\bm{A}^T \bm{A})^{-1} \bm{A}^T \bm{y}$. Matrix decomposition methods, especially SVD, are efficient in computing $(\bm{A}^T \bm{A})^{-1} \bm{A}^T$.

A fundamental prerequisite of conventional ELM is that all training data and corresponding labels must be available in advance. Consequently, the model requires retraining whenever new data and labels are available.

\subsection{OS-ELM}

Based on ELM, OS-ELM~\cite{huang2005line} can perform sequential training (batches of training data) instead of training all at once. Given the total dataset is chunked into $k_i$ batches of samples, where a chunk has $\bm{x}_i \in \mathbb{R}^{k_i \times m}$ and $\bm{y}_i \in \mathbb{R}^{k_i \times n}$, with $k$ being the batch size, learning $\bm{\eta}$ for the $i\textsuperscript{th}$ batch of samples can be expressed as

\begin{equation}
\norm{
\begin{bmatrix}
\bm{H}_0 \\
\vdots \\
\bm{H}_i
\end{bmatrix}
\bm{\eta}_i - 
\begin{bmatrix}
\bm{y}_0 \\
\vdots \\
\bm{y}_i
\end{bmatrix}
},
\label{eq:residual}
\end{equation}

where $\bm{H}_i = \varPhi(\bm{W} \bm{x}_i + \bm{b})$. $\bm{\eta}_i$ can be resolved by computing:
\begin{equation}
\begin{aligned}
\bm{P}_i &= \bm{P}_{i-1} - \bm{P}_{i-1} \bm{H}_i^T (\bm{I} + \bm{H}_i \bm{P}_{i-1} \bm{H}_i^T)^{-1} \bm{H}_i \bm{P}_{i-1}, \\
\bm{\eta}_i &= \bm{\eta}_{i-1} + \bm{P}_i \bm{H}_i^T (\bm{y}_i - \bm{H}_i \bm{\eta}_{i-1}),
\end{aligned}
\label{eq:recursive-update}
\end{equation}

where \( \bm{I} \) is the identity matrix. $\bm{P}_0$ and $\bm{\eta}_0$ can be computed as:
\begin{equation}
\begin{aligned}
\bm{P}_0 &= (\bm{H}_0^T \bm{H}_0)^{-1}, \\
\bm{\eta}_0 &= \bm{P}_0 \bm{H}_0^T \bm{y}_0.
\end{aligned}
\label{eq:initial-estimate}
\end{equation}

The number of training samples in the initial batch needs to be greater than $L$ to ensure that $\bm{H}_0^T \bm{H}_0$ is not singular. A previous study~\cite{tsukada2020neural} reveals that the matrix inversion in Eq.~\eqref{eq:recursive-update} can be converted to a scalar number inversion by making the sequential batch size $k = 1$, since the size of $\bm{I} + \bm{H}_i \bm{P}_{i-1} \bm{H}_i^T$ is $1 \times 1$. Therefore, Eq.~\eqref{eq:recursive-update} can be simplified as follows:

\begin{equation}
\begin{aligned}
\bm{P}_i &= \bm{P}_{i-1} - \frac{\bm{P}_{i-1} \bm{h}_i^T \bm{h}_i \bm{P}_{i-1}}{1 + \bm{h}_i \bm{P}_{i-1} \bm{h}_i^T}, \\
\bm{\eta}_i &= \bm{\eta}_{i-1} + \bm{P}_i \bm{h}_i^T (\bm{y}_i - \bm{h}_i \bm{\eta}_{i-1}),
\end{aligned}
\label{eq:rank1-update}
\end{equation}

where $\bm{h} \in \mathbb{R}^L$ is equivalent to $\bm{H} \in \mathbb{R}^{k \times L}$ when $k = 1$. Therefore, the hardware-expensive matrix inversion operation is eliminated, as the dimension of $1 + \bm{h}_i \bm{P}_{i-1} \bm{h}_i^T$ in Eq.~\eqref{eq:rank1-update} is $1 \times 1$, which simplifies the hardware implementation afterwards.

However, IT is still a notable computing overhead, as the size of $\bm{H}_0^T \bm{H}_0$ is $k \times k$, whose inversion also incurs significant hardware consumption and latency. Therefore, a hardware-efficient matrix decomposition is critical to computing IT before OBT. As we aim to implement an improved OS-ELM on SoC-based FPGA to take advantage of both multicore CPU and programmable logic, we propose to integrate OJR-SVD into OS-ELM, as OJR-SVD has been validated for low latency and improved accuracy even for large matrices on embedded devices~\cite{alessandrini2020singular}.

\subsection{OSOS-ELM}

\begin{algorithm}[t!] 
    \caption{Pseudo-Algorithm for OSOS-ELM}
    \begin{algorithmic}[1]
        \State $\bm{H}_0 = \mathrm{\varPhi}(\bm{W} \bm{x}_0 + \bm{b})$
        \State Compute $\bm{A} = \bm{H}_0^T \bm{H}_0$
        \State OJR-SVD on $\bm{A}$: $[\bm{U}, \bm{S}, \bm{V}] = \textit{OJR\_SVD}(\bm{A})$
        \State Transpose $\bm{S}$: $\bm{S} = \bm{S}^T$
        \State Calculate tolerance: $\textit{tol} = \max(\text{size}(\bm{A})) \times \epsilon(\|\bm{S}^T\|_{\infty})$ \Comment{$\epsilon$: gap between consecutive floating-point numbers}
        \State Determine rank threshold: $r_1 = \sum(\bm{S} > \textit{tol}) + 1$
        \State Truncate $\bm{U}$, $\bm{V}$, and $\bm{S}$ at rank $r_1$
        \State \hspace{1em} $\bm{V}(:, r_1:\text{end}) \gets []$
        \State \hspace{1em} $\bm{U}(:, r_1:\text{end}) \gets []$
        \State \hspace{1em} $\bm{S}(r_1:\text{end}) \gets []$
        \State Compute reciprocal of singular values: $\bm{s} = 1 ./ \bm{S}(:)$
        \State Compute pseudo-inverse: $\bm{P}_{N_0} = \bm{V} \bm{s}^T \bm{U}^T$
        \State OJR-SVD on $\bm{H}_0$: $[\bm{U}_b, \bm{S}_b, \bm{V}_b] = \textit{OJR\_SVD}(\bm{H}_0)$
        \State Transpose $\bm{S}_b$: $\bm{s}_b = \bm{S}_b^T$
        \State Calculate tolerance: $\textit{tol} = \max(\text{size}(\bm{H}_0)) \times \epsilon(\|\bm{S}_b\|_{\infty})$
        \State Determine rank threshold: $r_1 = \sum(\bm{S}_b > \textit{tol}) + 1$
        \State Truncate $\bm{U}_b$, $\bm{V}_b$, and $\bm{S}_b$ at rank $r_1$
        \State \hspace{1em} $\bm{V}_b(:, r_1:\text{end}) \gets []$
        \State \hspace{1em} $\bm{U}_b(:, r_1:\text{end}) \gets []$
        \State \hspace{1em} $\bm{S}_b(r_1:\text{end}) \gets []$
        \State Compute reciprocal of singular values: $\bm{s}_b = 1 ./ \bm{s}_b(:)$
        \State Compute pseudo-inverse: $\bm{\eta}_{N_0} = \bm{V}_b \bm{s}_b^T \bm{U}_b^T$
        \State Multiply with $\bm{y}_{N_0}$: $\bm{\eta}_{N_0} = \bm{\eta}_{N_0} \cdot \bm{y}_{N_0}$
        \For{$i = N_0 + 1$ to $I$}
            \State $\bm{x}_i = \bm{x}(i{:}i+1,:)$
            \State $\bm{y}_i = \bm{y}(i{:}i+1,:)$
            \State $\bm{h} = \mathrm{\Phi}(\bm{W} \bm{x}_i + \bm{b})$
            \State $\bm{P}_i = \bm{P}_{i-1} - \frac{\bm{P}_{i-1} \bm{h}^T \bm{h} \bm{P}_{i-1}}{1 + \bm{h} \bm{P}_{i-1} \bm{h}^T}$
            \State $\bm{\eta}_i = \bm{\eta}_{i-1} + \bm{P}_i \bm{h}^T (\bm{y}_i - \bm{h} \bm{\eta}_{i-1})$
        \EndFor
        \If{inference}
            \State $\bm{H}_{\text{test}} = \mathrm{\Phi}(\bm{W} \bm{x}_{\text{test}} + \bm{b})$
            \State $\hat{\bm{y}} = \bm{H}_{\text{test}} \cdot \bm{\eta}$
        \EndIf
    \end{algorithmic}
    \label{alg:osos-elm}
\end{algorithm}

\textbf{Algorithm~\ref{alg:osos-elm}} presents the detailed operations of OSOS-ELM. The most crucial and computationally intensive operations are the pseudo matrix inverse of $\bm{H_0}$ and $\bm{H_0^T H_0}$, resolved by \textit{OJR\_SVD} (Lines $\textbf{3}$ and $\textbf{10}$). IT is described from Line \textbf{1} to \textbf{17}; one-batch training (OBT), which involves massive matrix multiplications, is described from Lines $\textbf{18}$ to $\textbf{26}$. We implemented \textit{OJR\_SVD} according to the original study ~\cite{demmel1992jacobi}.

Briefly, it requires $\bm{A} \in \mathbb{R}^{m \times n}$ ($m \geq n$), and a counter terminating the iteration and controlling the accuracy. $\bm{S}$, $\bm{U}$, and $\bm{V}$ denote the singular values, left singular vector, and right singular vector in $\bm{A}$. Two nested for-loops iterate through columns in $\bm{U}$, $\bm{V}$ to compute Jacobi rotation. The iterations are terminated when \textit{ counts} $\geq$ 15. The Jacobi method is preferable for our OSOS-ELM. It is faster for decomposing small matrices than QRD~\cite{takata2021implementation}, and in our case, the initial batch in IT is adjustable and could be a small matrix.

\textit{count} is a hyper-parameter and configurable in our hardware discussed thereafter, achieving the trade-off between accuracy and latency. We tested it from 1 to 20 during training for FLIM and DCS datasets and found that the accuracy does not increase when \textit{ counts} $>$ 15. The training process consists mainly of IT and OBT, spanning lines $\textbf{3}$ to $\textbf{23}$ and Lines $\textbf{24}$ to $\textbf{30}$, respectively.

In particular, the two \textit{OJR\_SVD} operations in IT can be executed independently, as there is no data dependency between them. These tasks are implemented separately on different CPU cores using a pseudo-interruption design on a standalone PS, as discussed in the hardware implementation hereafter. Meanwhile, OBT relies on the results from IT and involves extensive matrix and vector operations, making it well-suited for implementation on the PL.

\subsection{Datasets Description of FLIM, DCS, and LiDAR}
\label{sec:Datasets}
As the modeling of fluorescence decays and ACFs is well studied~\cite{xiao2021dynamic, smith2019fast, poon2020deep, wang2024quantification}, we briefly introduce the essential concept of mathematical models. Fluorescence lifetime reveals cellular metabolism, protein interactions, and tumor microenvironments in biology and medicine. $\tau$ can be reconstructed from single- or multi-exponential functions acquired and encoded by FLIM systems~\cite{becker2012fluorescence}. $\tau$ is the variable to be retrieved from the decay that can be modeled as

\begin{equation}
h(t) = IRF(t) * P \sum_{k=1}^{K} \alpha_k e^{-t/\tau_k} + n(t),
\label{eq:fluorescence-decay-model}
\end{equation}

where $IRF(\cdot)$ is the system’s instrument response function, $P$ represents the amplitude, $\tau_k$ is the $k^{\text{th}}$ lifetime component, $\alpha_k$ is the $k^{\text{th}}$ amplitude fraction, and $n(t)$ includes Poisson noise~\cite{fereidouni2017rapid} and dark count rate of the sensor. $t = [1, 2, \dots, T]$ is the time-bin index of the TCSPC module. Therefore, $h(t)$ composed of ranges of lifetimes and optical parameters for bespoke experimental setups and applications can be generated. As bi-exponential models ($K=2$) can well estimate multi-exponential decays~\cite{li2020investigations}, $\tau_I$ and $\tau_A$ are employed for quantifying Förster resonance energy transfer and fluorescence quenching behaviors. Therefore, in our case, we take $\tau_I$ and $\tau_A$ as the outputs of the OSOS-ELM model.

DCS detects dynamic speckles at a single-photon detector, where speckles are generated by the fluctuating intensity of red blood cells in live tissue. ACFs can be computed by measuring intensity images $I(t)$ at each time step,

\begin{equation}
g_2(\tau_{DCS}) = \frac{\langle I(t) \cdot I(t + \tau_{DCS}) \rangle}{\langle I(t) \rangle^2},
\label{eq:acf-dcs}
\end{equation}

where $\tau_{DCS}$ is the time lag, and $\langle \cdot \rangle$ averages the intensity of detected photons over time. The unnormalized electric field correlation function $G_1(\tau_{DCS})$ is represented by the correlation diffusion equation~\cite{boas1997spatially}. $G_1(\tau_{DCS})$ is computed using geometry parameters under the extrapolated boundary condition and the continuous-wave condition. $G_1(\tau_{DCS})$ is the solution for the correlation diffusion equation and includes $\alpha D_B$ (equivalent to BFi), where $\alpha$ is the ratio of moving scatters to all scatters, $D_B$ is the effective Brownian diffusion coefficient of scatters.

Given $G_1(\tau_{DCS})$, an ACF $g_2(\tau_{DCS})$ can be computed by the Siegert relationship~\cite{rice1944mathematical}:
\begin{equation}
g_2(\tau_{DCS}) = 1 + \beta |g_1(\tau_{DCS})|^2; \quad g_1(\tau_{DCS}) = \left[ \frac{G_1(\tau_{DCS})}{G_1(\tau_{DCS=0})} \right],
\label{eq:siegert-relation}
\end{equation}
where $\beta$ is the coherence factor and is inversely proportional to the number of detected speckles. We refer to~\cite{helton2022numerical, dong2012simultaneously} to apply noise to generated clean $g_2(\tau_{DCS})$. We take BFi and $\beta$ as the outputs of the OSOS-ELM.

\begin{figure*}[t]
    \centering
    \includegraphics[width=\linewidth]{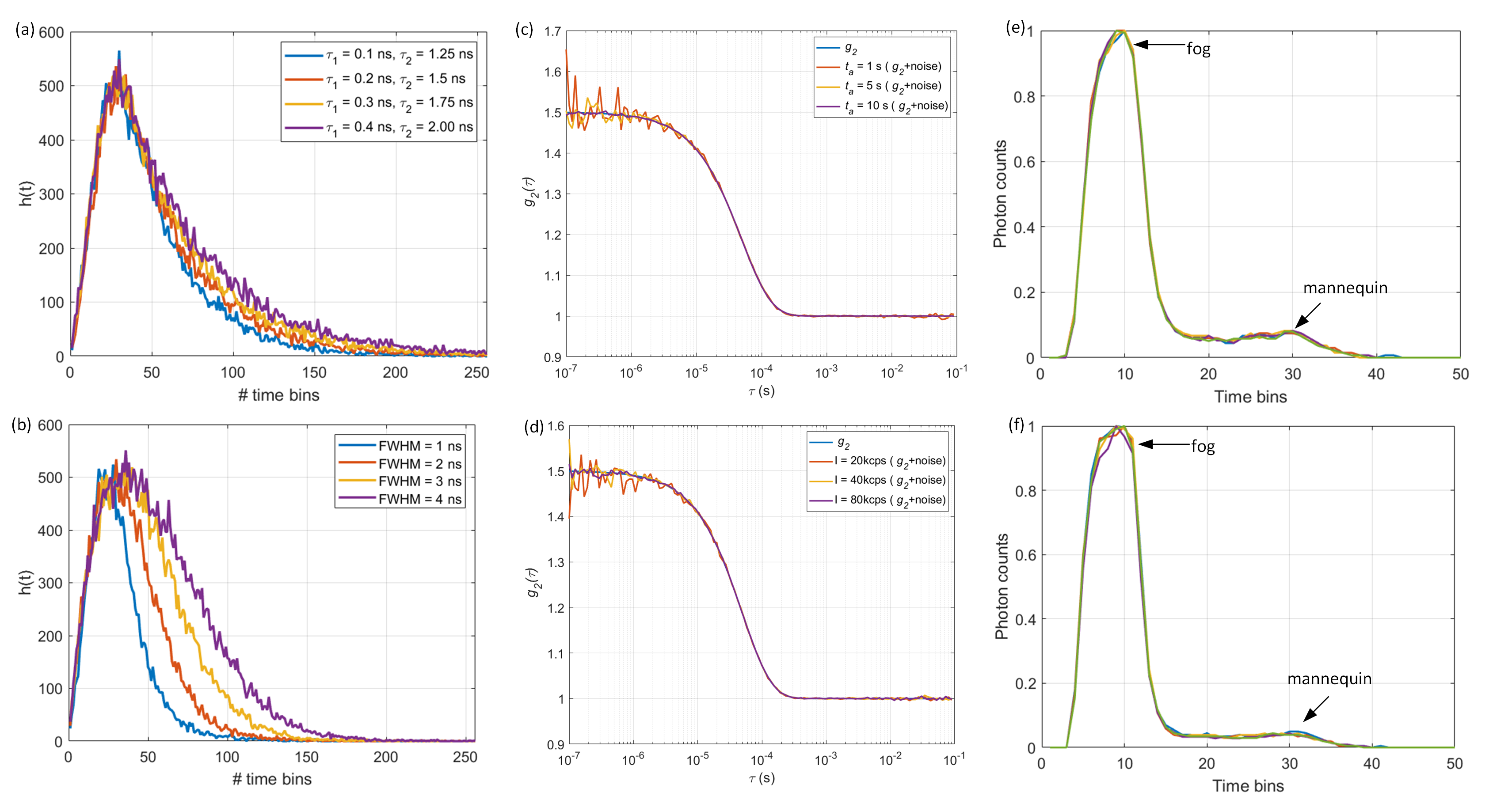}
    \caption{Examples of FLIM decays with (a) two lifetime components and (b) different laser FWHM. ACFs (clean and noise-applied) were generated using the same absorption and scattering coefficients and source-detector distance but with (c) different total averaging times ($t_a$) and (d) varying photon intensity. (e) and (f) also show four example histograms with normalized photon counts collected from a mannequin in a chamber with heterogeneous fog, labeled as class 1 and class 2, respectively.}
    \label{fig2:data_example}
\end{figure*}

Given the mathematical models of $h(t)$ for FLIM and $g_2(\tau)$ for DCS, synthetic datasets can be generated for the network’s training. We generate fluorescence decays and ACFs by different parameters, and noise levels are depicted in Fig.~\ref{fig2:data_example}, covering the most common cases. Unlike DCS and FLIM, which use mathematical models to generate synthetic training datasets, dataset generation for sensing-through-fog LiDAR is challenging due to the complexity of modeling heterogeneous fog and the random walk of photons. Therefore, the LiDAR datasets used in this study were collected in our previous work~\cite{zang2024object}.

\section{IMPLEMENTATION DETAILS}
\subsection{Training Data Generation}
As discussed in Section~\ref{sec:Datasets}, there are two output nodes (\#ON) in our network for FLIM and DCS. The number of input nodes (\#IN) corresponds to the number of time bins in histograms and is set to 256, which is common for FLIM systems. The temporal resolution (histogram bin width) in Fig.~\ref{fig2:data_example}(a) and~(b) is $0.039~\text{ns}$. The ranges of the two lifetime components are $[0.1, 5]~\text{ns}$ and $[1, 3]~\text{ns}$, and the full width at half maximum (FWHM) is $0.1673~\text{ns}$. For DCS, two output nodes correspond to $\text{BFi}$ and $\beta$. Each ACF is composed of 128 non-linearly spaced lags (shown in Fig.~\ref{fig2:data_example}(c) and~(d)), ranging from $0.1~\text{s}$ to $10^{-7}~\text{s}$. The diffusion and absorption coefficients are set to $2~\text{mm}^{-1}$ and $0.1~\text{mm}^{-1}$, respectively. The illumination wavelength is $700~\text{nm}$, and the source-detector distance is $10~\text{mm}$. Further details on the preset parameters of DCS are provided in~\cite{wang2024quantification}. For the classification task of LiDAR, \#IN is 50, and \#ON is 1. The details of the datasets, along with the sensor and laser specifications, are presented in~\cite{zang2024object}.

\begin{figure}[t]
    \centering
    \includegraphics[width=\linewidth]{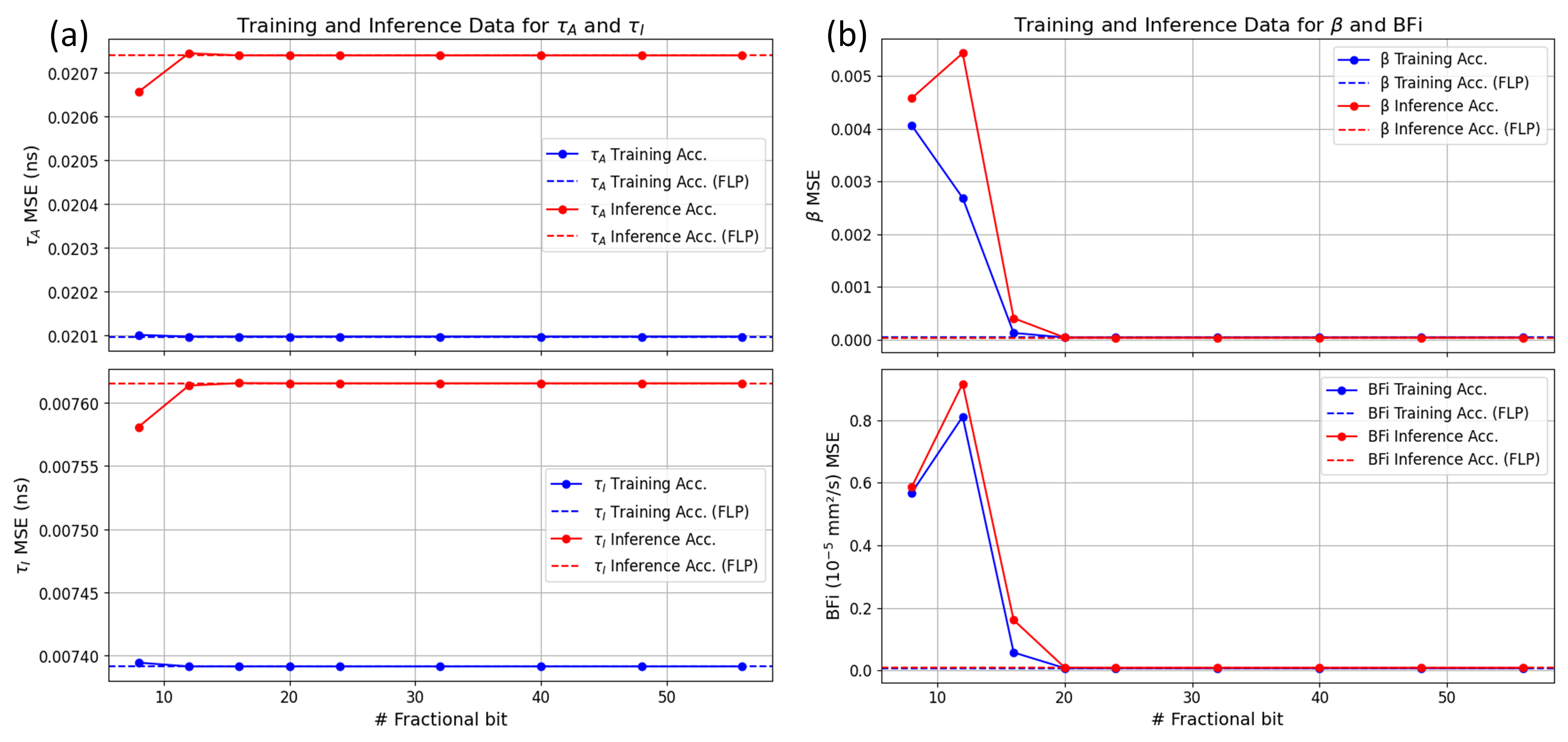} 
    \caption{Software evaluation of accuracy with different lengths of fractional bits in FXP format for (a) $\tau_A$ and $\tau_I$ of FLIM, and (b) BFi and $\beta$ of DCS. Blue and red dashed lines indicate the reference accuracy obtained by FLP.}
    \label{fig:figure3_accuracy_fxp}
\end{figure}

\subsection{FPGA Implementation}
Although fixed-point arithmetic (FXP) can save hardware resources and reduce computational latency, our applications are sensitive to rounding errors—especially for the vector of singular values (at Line~$\textbf{3}$ and $\textbf{13}$ in Algorithm~\ref{alg:osos-elm}, precise to ten decimal places)—which can deteriorate subsequent computations due to propagation errors. Additional hardware logic is also required to handle overflow and underflow. As the computing in the processing system (PS) is floating-point (FLP)-based, converting from FLP to FXP (using division or bit-shift operations) for the programmable logic (PL) induces extra computational overhead on the PS. Apart from computational workload, we also evaluate the accuracy of different lengths of fractional bits of FXP for training and inference in FLIM and DCS. As shown in Fig.~\ref{fig:figure3_accuracy_fxp}, we use the \texttt{quantizer} function in MATLAB to simulate FXP operations of OBT on hardware with overflow and underflow corrections, providing a reference for hardware implementation. The dashed lines represent the reference accuracy from FLP simulation in MATLAB. Fig.~\ref{fig:figure3_accuracy_fxp}(a) indicates that the accuracy of lifetime reconstruction is robust to the number of fractional bits, aligning with the mean squared error (MSE) of FLP, except when the number of fractional bits is set to 8. However, as shown in Fig.~\ref{fig:figure3_accuracy_fxp}(b), the MSE of DCS converges when the number of fractional bits is greater than 20 (equivalent to a precision of $1/2^{20} = 9.54 \times 10^{-7}$), because the analytical model of DCS is more complex than FLIM. This complexity makes the training and inference stages more sensitive to rounding errors. Therefore, the number of fractional bits serves as a hyper-parameter in our design and significantly affects generalization. Considering the trade-off between computational overhead and accuracy, we used double-precision FLP for software simulations, aligning with the implementations in both MATLAB and hardware. We also investigated the effect of FXP in sensing-through-fog LiDAR. The resulting classification accuracy remains consistent despite variations in the number of fractional bits. This is because classification results are discrete integers, unlike continuous-valued regression tasks. Hence, classification tasks are more tolerant of numerical precision variations. The IT and OBT are implemented across the PS and PL of a ZCU104 Ultrascale+ MPSoC FPGA. We verified that the outputs from both FPGA and CPU implementations are identical. Finally, we demonstrate the hardware utilization, latency, and power consumption of OSOS-ELM on the FPGA and compare them with those on a GPU platform in Section~\ref{sec:hardware-eval}.

\begin{table}[t]
    \caption{Training Latency of Each Operation in OBT}
    \label{tab:latency_obt}
    \centering
    \renewcommand{\arraystretch}{1.1}
    \resizebox{\columnwidth}{!}{%
    \begin{tabular}{|l|l|c|c|}
        \hline
        \textbf{Operations} & \textbf{Expression} & \textbf{DCS (cc)} & \textbf{FLIM (cc)} \\
        \hline
        \multirow{5}{*}{MVM} 
        & $\bm{h}_i = \varPhi(\bm{W} \bm{x}_i + \bm{b})$ & 19,913 & 199,800 \\
        & $\bm{c} = \bm{P}_{i-1} \bm{h}_i^T$ & 11,934 & 11,934 \\
        & $\bm{a} = (1 + \bm{h}_i \bm{P}_{i-1} \bm{h}_i^T)^{-1}$ & 11,943 & 11,943 \\
        & $\bm{h}\bm{\eta} = \bm{h}_i \cdot \bm{\eta}_{i-1}$ & 1,530 & 1,530 \\
        & $\bm{Py\_}h\bm{\eta} = \bm{P}_i \cdot \bm{h}_i^T (\bm{y}_i - \bm{h}\bm{\eta})$ & 372 & 372 \\
        \hline
        MMM & $\bm{c} \cdot \bm{a} \cdot \bm{d}^1$ & 22,517 & 22,517 \\
        \hline
        \multirow{3}{*}{V/M add/sub} 
        & $\bm{P}_i = \bm{P}_{i-1} - \bm{c} \cdot \bm{a} \cdot \bm{d}$ & 22,524 & 22,524 \\
        & $\bm{y}\_h\bm{\eta} = \bm{y}_i - \bm{h}\bm{\eta}$ & 75 & 75 \\
        & $\bm{\eta}_{i-1} + \bm{Py\_}h\bm{\eta}$ & 306 & 306 \\
        \hline
        \multicolumn{2}{|l|}{$\bm{W}$ and $\bm{b}$ loading} & 45,002 & 45,002 \\
        \hline
        \multicolumn{2}{|l|}{Matrix initialization/update} & 13,967 & 13,967 \\
        \hline
    \end{tabular}
    }
    \vspace{1mm}
    \begin{flushleft}
    \footnotesize%
        $^1\ \bm{d} = \bm{h}_i \bm{P}_{i-1}$, included in $\bm{a}$.
    \end{flushleft}
\end{table}

\begin{table}[t]
    \caption{Inference Latency of OSOS-ELM}
    \label{tab:inference_latency}
    \centering
    \renewcommand{\arraystretch}{1.1}
    \resizebox{\columnwidth}{!}{%
    \begin{tabular}{|l|l|c|c|}
        \hline
        \textbf{Operations} & \textbf{Expression} & \textbf{DCS (cc)} & \textbf{FLIM (cc)} \\
        \hline
        \multirow{2}{*}{MVM} 
        & $\bm{H}_{\text{test}} = \varPhi(\bm{W} \bm{x}_{\text{test}} + \bm{b})$ & 9,663 & 19,295 \\
        & $\bm{\hat{y}} = \bm{H}_{\text{test}} \cdot \bm{\eta}$ & 379 & 379 \\
        \hline
        \multicolumn{2}{|l|}{$\bm{W}$ and $\bm{b}$ loading} & 45,002 & 45,002 \\
        \hline
        \multicolumn{2}{|l|}{Matrix initialization/update} & 150 & 150 \\
        \hline
    \end{tabular}
    }
    \vspace{1mm}
\end{table}

\begin{figure*}[t]
    \centering
    \includegraphics[width=\linewidth]{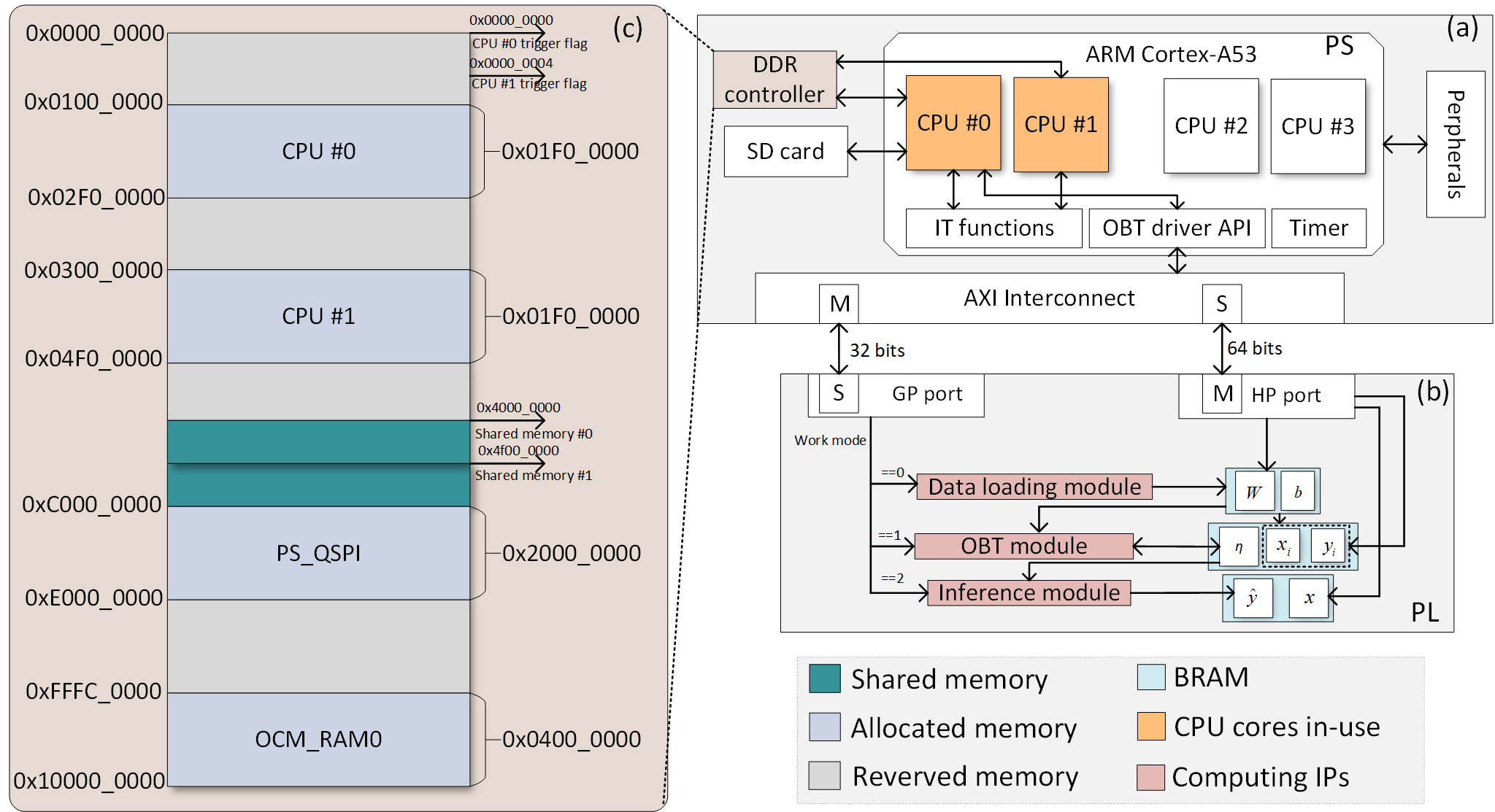} 
    \caption{Overview of the hardware architecture. (a) Enabled CPU cores (orange), functions computing IT, and instantiated hardware APIs in PS; . Three IP cores, i.e., data loading module, training module, and inference module; (c) Memory segmentation and addresses for storing matrices and flag signals.}
    \label{fig:figure4_HW_overview}
\end{figure*}

\subsubsection*{IT Implementation on PS}
IT is implemented on the PS using a bare-metal platform, which is more suitable for time-critical computing than \textit{PetaLinux} due to reduced memory management overhead and OS-induced latency. Additionally, SPAD-coupled FPGAs operate without an OS. Thanks to the inherent parallelism potential of two independent \textit{OJR\_SVD} functions described in Algorithm~\ref{alg:osos-elm}, the IT computation from Lines~$\textbf{2}$ to $\textbf{12}$ and from Lines~$\textbf{13}$ to $\textbf{23}$ in Algorithm~\ref{alg:osos-elm} is split and deployed on two CPU cores for multi-threading, since $\bm{H_0}$ is the only input for both of these independent tasks. There are four ARM Cortex-A53 CPU cores available, and we allocate CPU\#0 and CPU\#1 for this multi-threading application. As shown in Fig.~\ref{fig:figure4_HW_overview}(c), the memory is segmented into two individual spaces (starting at \texttt{0x0100\_0000} and \texttt{0x0300\_0000}, respectively) for CPU\#0 and CPU\#1, to ensure isolated memory access and avoid memory overlap that could lead to faults. Two addresses (\texttt{0x0000\_0000} and \texttt{0x0000\_0004}) are configured as trigger flags for their corresponding CPU cores. According to Algorithm~\ref{alg:osos-elm}, once $\bm{H_0}$ is obtained from Line~$\textbf{1}$ in CPU\#0, the trigger flag (located at address \texttt{0x0000\_0004}) is flipped from 0 to 1, triggering CPU\#1 to begin its task at Line~$\textbf{13}$. Matrix $\bm{H_0}$ is then sent to shared memory at address \texttt{0x4000\_0000}. CPU\#1 continuously polls this address until the flag is captured, then reads $\bm{H_0}$ and begins computing \textit{OJR\_SVD}($\bm{H_0}$) at Line~$\textbf{13}$. Simultaneously, CPU\#0 computes \textit{OJR\_SVD}($\bm{A}$) at Line~$\textbf{3}$ using the same shared memory. When CPU\#1 finishes computing from Lines~$\textbf{13}$ to $\textbf{23}$, the next stage of processing is triggered. The resultant matrix $\bm{\eta}$ is loaded into another shared memory (\texttt{0x4f00\_0000}), and the trigger signal for CPU\#0 is flipped to notify CPU\#0 to fetch $\bm{\eta}$. At this point, all data is back to CPU\#0 and ready for OBT on PL (from Lines $\textbf{24}$ to $\textbf{30}$).

After IT finishes, CPU\#0 uses OBT driver APIs to change the working mode and invoke the IP cores. A physical timer counter of CPU\#0 was used to measure the latency of IT and OBT. The optimization flag \texttt{-O1} is applied to optimize the code compilation and accelerate IT. Notably, accessing the triggering flags for CPU\#0 and \#1 is a pseudo-interruption mechanism, saving the overhead of explicitly executing interruption API calls. To mimic real data acquisition and storage scenarios, training data and labels of IT are pre-stored on hardware. We select the SD card to store the data because of its large capacity and acceptable latency for real-time data fetching. Onboard DDR can also be used, providing higher bandwidth but with smaller capacity. 
Afterward, the data is read from the SD card to the onboard DDR for IT. Similarly, training data and labels of OBT follow the same data-fetching pipeline, with the only difference being that OBT operates via batch-by-batch reading.

\subsubsection*{OBT Implementation on PL}
OBT is implemented using \textit{Vivado HLS 2018.3}. As shown in Fig.~\ref{fig:figure4_HW_overview}(c), there are three distinct IP cores implemented in PL. Before OBT (Line~$\textbf{24}$ in Algorithm~\ref{alg:osos-elm}), by setting \texttt{working mode} flag to 0, \textit{data loading module} preloads the parameters $\bm{W}$ and $\bm{b}$ that are initialized by PS to on-chip memory on PL. Afterward, \texttt{working mode} is set to 1 for OBT. Prefetched parameters $\bm{W}$ and $\bm{b}$ are dispatched to the \textit{training module} while training samples and labels are input to the module. To improve the throughput of the hardware, $\bm{W}$ and $\bm{b}$ are dispatched from BRAMs to register files for parallel access. Once the index of the input training sample reaches $\bm{I}$, OBT finishes, and $\bm{\eta}_{N0}$ is stored in BRAMs. After OBT finishes, \texttt{working mode} can be set to 2 for inference, and BRAM holding $\bm{\eta}_{N0}$ is ready for processing the new input sample $\bm{x}$. OBT (from Lines~$\textbf{24}$ to $\textbf{30}$ in Algorithm~\ref{alg:osos-elm}) mainly involves matrix-vector multiplication (MVM), matrix-matrix multiplication (MMM), and scalar division. Unrolling and pipelining optimizations are applied to these matrix- or vector-based operations. After passing C and RTL simulations, the hardware instance is exported as an IP core for system integration. The IP core is wrapped with AXI-full and AXI-lite interfaces to transfer matrices and configuration parameters, respectively, between CPU cores and the IP core. OBT training and inference on PL are implemented with distinct logic, without reuse. Although reusing the hardware of MVM and MMM can save hardware utilization, it prevents simultaneous training and inference. We reserve the potential for concurrent computing of training and inference when required. A multiplexer distinguishes the signal transferred from PS through AXI-lite to configure the \texttt{working mode}. 

The most computationally intensive operation in OBT is 
\begin{equation}
\bm{P}_i = \bm{P}_{i-1} - \frac{\bm{P}_{i-1} \bm{h}_i^T \bm{h}_i \bm{P}_{i-1}}{1 + \bm{h}_i \bm{P}_{i-1} \bm{h}_i^T},
\label{eq:Pi_update}
\end{equation}
where $\bm{P}_{i-1} \bm{h}_i^T$ and $\bm{h}_i \bm{P}_{i-1}$ are independent and implemented with a parallelized MVM module. 
$\bm{P}_{i-1} \bm{h}_i^T \bm{h}_i \bm{P}_{i-1}$ is the only MMM in the OBT for-loop. The dimensions of the operations are:
\begin{align*}
    &\bm{P}_{i-1} \in \mathbb{R}^{L \times L}, 
    \quad \bm{h}_i^T \in \mathbb{R}^{L \times 1}, 
    \quad \bm{h}_i \in \mathbb{R}^{1 \times L}, \\
    &\bm{h}_i \bm{P}_{i-1} \bm{h}_i^T \in \mathbb{R}^{1 \times 1}, 
    \quad \text{so the full update term is in } \mathbb{R}^{L \times L}.
\end{align*}

The breakdown of training and inference latency in clock cycles (cc) for each MVM and MMM in DCS (\#IN = 128, $L = 150$, \#ON = 2) and FLIM (\#IN = 256, $L = 150$, \#ON = 2) is shown in \textbf{Table~\ref{tab:latency_obt}} and \textbf{Table~\ref{tab:inference_latency}}, respectively. 

Operations in $\varPhi(\bm{W} \bm{x} + \bm{b})$ can be merged into two nested \texttt{for}-loops. Other operations, especially in Eq.~\eqref{eq:Pi_update}, are unrolled and pipelined to minimize latency. Latency arises during matrix initialization or merging, as intermediate results must be reset to zero between training iterations. With an identical unrolling strategy and the same $L$ and \#ON, the only difference in latency between DCS and FLIM stems from $\varPhi(\bm{W} \bm{x})$, since the dimensions of $\bm{W}$ and $\bm{x}$ are tied to \#IN. Likewise, as shown in \textbf{Table~\ref{tab:inference_latency}}, the total latency of inference is determined by $\varPhi(\bm{W} \bm{x}_{\text{test}} + \bm{b})$.

Once the \texttt{working mode} is switched from training mode to inference mode, inference will be executed (from Lines $\textbf{31}$ to $\textbf{34}$). 

\begin{figure*}[t]
    \centering
    \includegraphics[width=\linewidth]{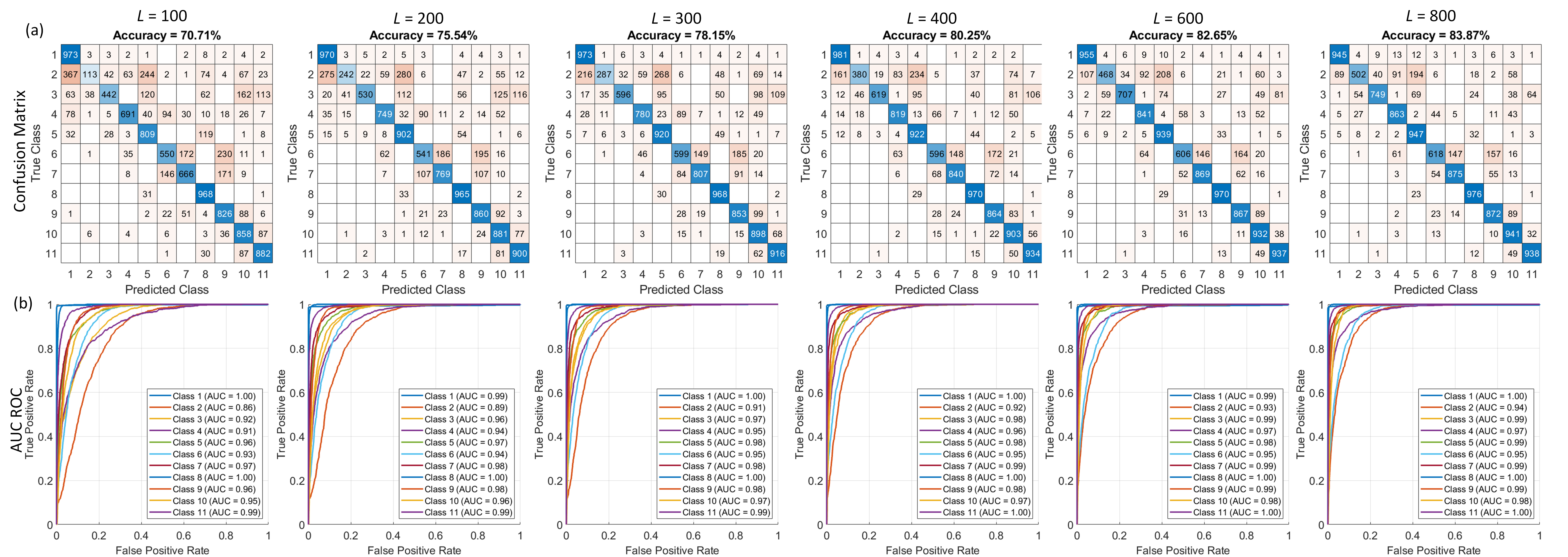} 
    \caption{Evaluation of object classification using LiDAR histograms from OSOS-ELM. (a) Confusion matrices and average accuracy. (b) AUC scores and ROC curves for each class.}
    \label{fig:figure5_accuracy_lidar}
\end{figure*}

\begin{figure}[t]
    \centering
    \includegraphics[width=\linewidth]{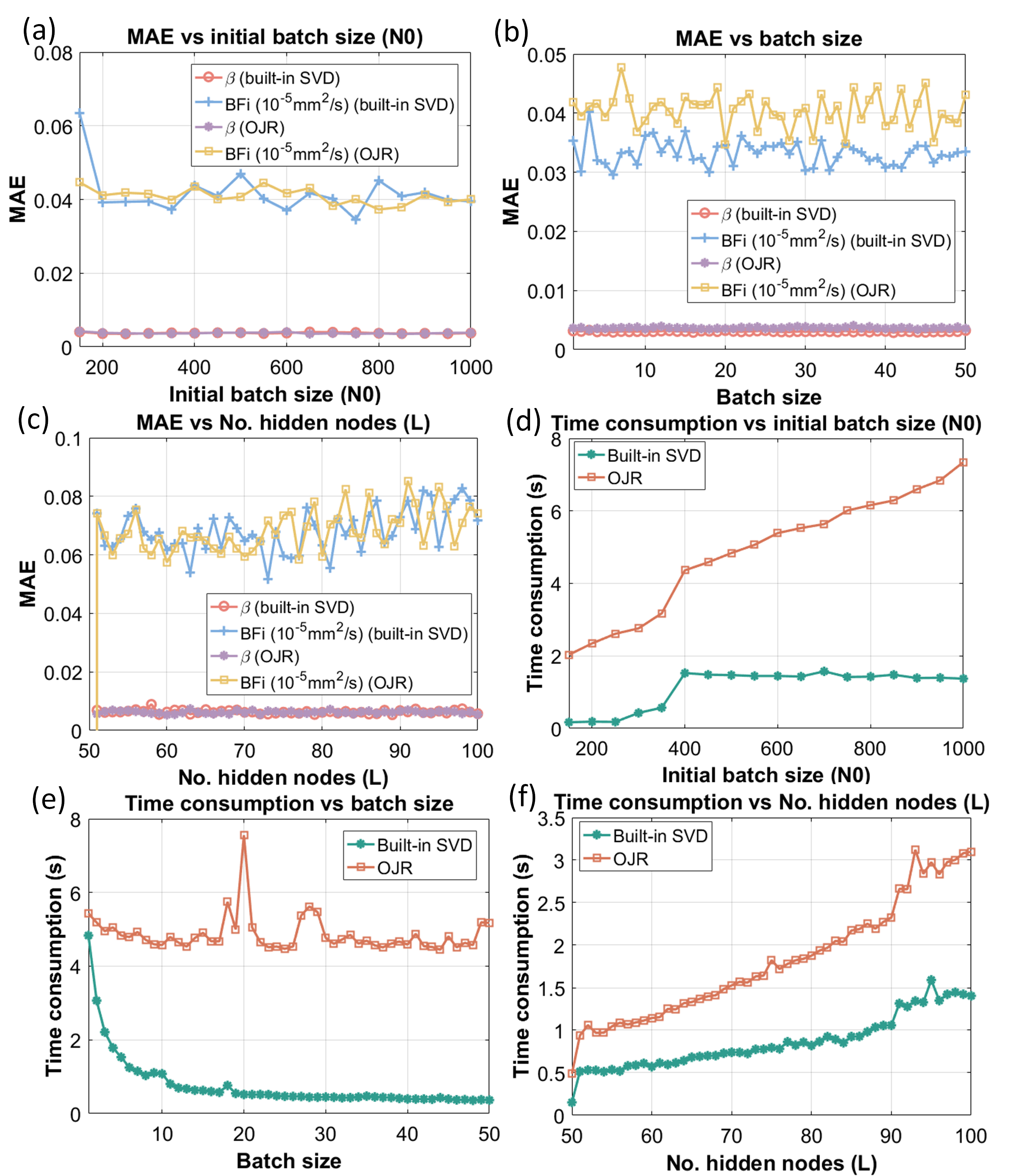} 
    \caption{Evaluation of OJR-SVD-based and MATLAB built-in SVD-based training for (a)--(f) BFi and $\beta$ reconstruction for DCS. Batch size, $L$, and $N_0$ are variables to investigate the effect of time consumption and MAE.}
    \label{fig:figure6_accuracy_dcs}
\end{figure}

\begin{figure}[t]
    \centering
    \includegraphics[width=\linewidth]{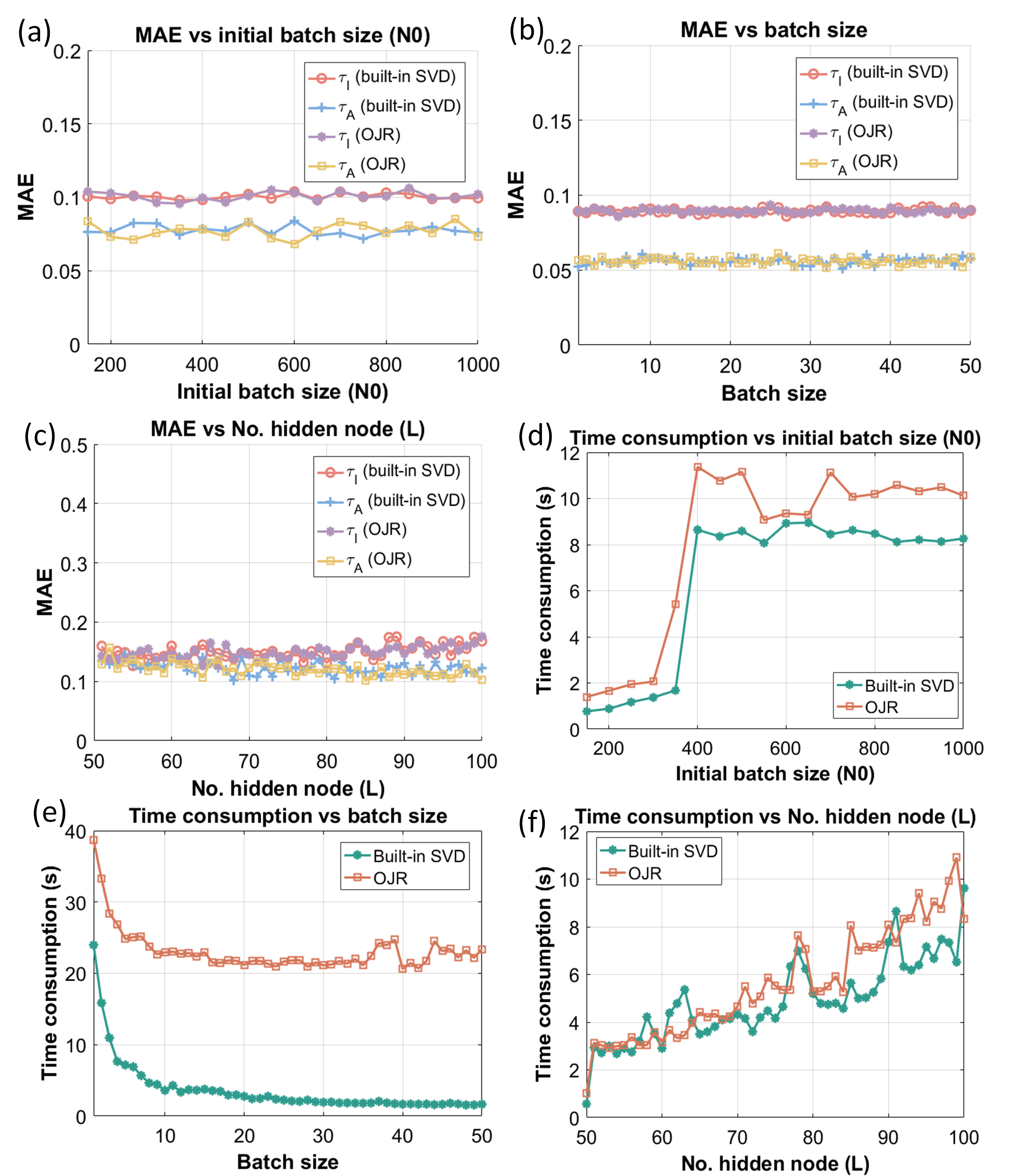} 
    \caption{Evaluation of OJR-SVD-based and MATLAB built-in SVD-based training for (a)--(f) fluorescence lifetime reconstruction for FLIM. Batch size, $L$, and $N_0$ are variables to investigate the effect of time consumption and MAE.}
    \label{fig:figure7_accuracy_flim}
\end{figure}

\subsection{GPU Comparison}
To comprehensively evaluate the computational efficiency of OSOS-ELM, we benchmark its performance across heterogeneous platforms, focusing on latency and power consumption. For a fair comparison, we use the Jetson Xavier NX, a \textit{low-power} yet high-performance embedded edge AI device developed by NVIDIA, as our system emphasizes compactness and power efficiency. The system has a 6-core ARM Cortex-A76 CPU and a 384-core NVIDIA Volta GPU with Tensor Cores for AI acceleration. The device is set to a power configuration of 20W with 4 cores enabled for power efficiency. GPU kernels are programmed using CUDA~\cite{Nickolls2008Scalable} v11.4, while CPU tasks are coded in C++ and compiled with GCC 9.4. GPUs are optimized for highly parallel computations by emphasizing data processing instead of data caching and flow control. Therefore, although GPUs are powerful and highly efficient for parallelizable tasks, they struggle with sequential processing and can encounter inefficiencies when a lot of branching logic is present. For this reason, the \textit{OJR\_SVD} algorithm, which involves branching and many sequential operations, is left to be executed on the CPU. Similar to our FPGA-based approach, we employ a multi-thread strategy to parallelize the \textit{OJR\_SVD} computations, enabling two independent decompositions to run concurrently on two CPU cores.

\begin{figure}[t]
    \centering
    \includegraphics[width=\linewidth]{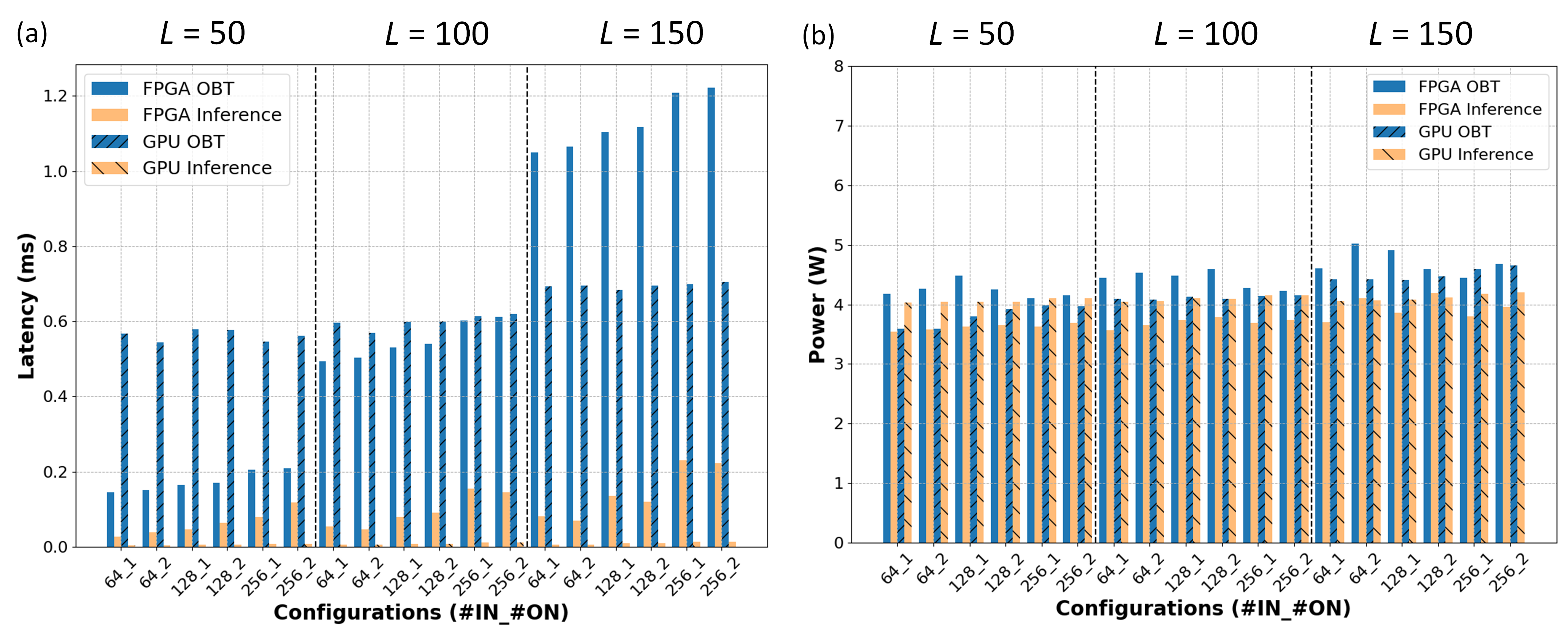} 
    \caption{Comparison of latency and power consumption measured from FPGA and GPU, at different network configurations and values of $L$.}
    \label{fig:figure8_GPU_compare}
\end{figure}

A key feature of NVIDIA's GPU architecture is the Streaming Multiprocessors (SMs), which support Single Instruction Multiple Thread (SIMT) execution. In other words, each SM allows concurrent execution of multiple threads, and there are multiple SMs per GPU. In our GPU implementation, we leverage SMs to improve overall performance by parallelizing independent tasks in OBT. The GPU implementation is benchmarked in terms of latency and power consumption versus the FPGA implementation. The evaluation is performed with different OSOS-ELM topologies to assess performance metrics under different computational loads. The number of input nodes \#IN ranges within \{64, 128, 256\}, the number of hidden nodes $L$ ranges within \{50, 100, 150\}, and the number of output nodes \#ON is chosen from \{1, 2\}. The latency is measured as the time elapsed from the moment the model receives an input until it completes either a prediction or an update of the output weights. All the reported times are averaged over 100{,}000 samples.

As for the power measurement, NVIDIA's Jetson platforms lack support for the standard system monitoring tool  \texttt{nvidia-smi}. However, an onboard INA3221 power monitor sensor on Jetson Xavier NX can be accessed via the  \texttt{tegrastats} utility or by Linux’s \texttt{sysfs} virtual file system. Instead of manually reading the data, for a more accurate power consumption measurement, we use a method similar to that in~\cite{holly2020profiling}, where a separate process is created to keep logging the voltage and current from the power monitor sensor. Simultaneously, this process launches the main function to be evaluated. The main function logs the start and end times of the OBT training and inference phases, which are then saved to a .txt file. After execution completes, the power and timing logs are analyzed, and the average power over the execution is computed. Unlike the FPGA platform, the GPU operates with a Linux operating system (OS). To minimize OS-related overhead, its graphical interface is disabled, and all control is performed via Secure Shell (SSH) with command-line interface (CLI).

\section{Performance Evaluation}
\label{sec:hardware-eval}
We first evaluate the classification accuracy of LiDAR in object recognition through fog. As shown in the confusion matrices in Fig.~\ref{fig:figure5_accuracy_lidar}(a), the average accuracy improves as $L$ increases. However, no significant accuracy enhancement is observed beyond $L = 600$. Receiver operating characteristic (ROC) curves and area under the curve (AUC) scores are also used to visualize classification performance. Like the confusion matrices, classes 2 and 6 present the most challenging cases. When evaluating OSOS-ELM with built-in SVD and OJR-SVD, we found that the classification results remained identical, indicating that the choice of matrix inversion method does not significantly impact performance. Additionally, given the large training dataset (990{,}000 histograms), increasing $N_0$ significantly increases the computational burden of IT. To mitigate this, we set $N_0 = 1000$ to balance computational efficiency and performance. Table~\ref{tab:lidar_runtime} presents the processing time for test datasets, showing that built-in SVD requires less time, as MATLAB efficiently handles large matrix decompositions. However, it is not scalable for FPGA implementation.

\begin{table*}[t]
\caption{Hardware consumption of training and inference for different dimensions of the model}
\centering
\renewcommand{\arraystretch}{1.2}
\begin{tabular}{|c|c|c|c|c|c|}
\hline
\#IN & \#ON & BRAM (Total: 312) & DSP (Total: 1,728) & FF (Total: 460,800) & LUT (Total: 230,400) \\
\hline
\multicolumn{6}{|c|}{\textbf{Hidden Nodes = 50}} \\
\hline
64  & 1 & 32.50 (T) + 24 (I) (18.10\%) & 56 (T) + 68 (I) (7.18\%) & 38,266 (T) + 13,208 (I) (11.17\%) & 31,921 (T) + 9,864 (I) (18.14\%) \\
64  & 2 & 32.50 (T) + 24 (I) (18.10\%) & 56 (T) + 71 (I) (7.35\%) & 38,497 (T) + 17,222 (I) (12.10\%) & 32,122 (T) + 13,041 (I) (19.60\%) \\
128 & 1 & 38.50 (T) + 35 (I) (23.56\%) & 56 (T) + 68 (I) (7.18\%) & 50,495 (T) + 17,790 (I) (14.82\%) & 37,307 (T) + 11,811 (I) (21.32\%) \\
128 & 2 & 38.50 (T) + 35 (I) (23.56\%) & 56 (T) + 71 (I) (7.35\%) & 50,723 (T) + 21,740 (I) (15.73\%) & 37,544 (T) + 14,980 (I) (22.80\%) \\
256 & 1 & 49.50 (T) + 56 (I) (33.81\%) & 56 (T) + 68 (I) (7.18\%) & 75,358 (T) + 25,982 (I) (21.99\%) & 49,554 (T) + 15,646 (I) (28.30\%) \\
256 & 2 & 49.50 (T) + 56 (I) (33.81\%) & 56 (T) + 71 (I) (7.35\%) & 75,633 (T) + 29,804 (I) (22.88\%) & 49,600 (T) + 18,793 (I) (29.68\%) \\
\hline
\multicolumn{6}{|c|}{\textbf{Hidden Nodes = 100}} \\
\hline
64  & 1 & 62.50 (T) + 35 (I) (31.25\%) & 56 (T) + 68 (I) (7.18\%) & 53,627 (T) + 13,226 (I) (14.51\%) & 43,656 (T) + 9,927 (I) (23.26\%) \\
64  & 2 & 62.50 (T) + 35 (I) (31.25\%) & 56 (T) + 71 (I) (7.35\%) & 53,874 (T) + 22,152 (I) (16.80\%) & 43,777 (T) + 14,876 (I) (25.46\%) \\
128 & 1 & 73.50 (T) + 56 (I) (41.50\%) & 56 (T) + 68 (I) (7.18\%) & 63,600 (T) + 17,809 (I) (17.67\%) & 49,810 (T) + 11,938 (I) (26.80\%) \\
128 & 2 & 73.50 (T) + 56 (I) (41.50\%) & 56 (T) + 71 (I) (7.35\%) & 63,757 (T) + 26,671 (I) (19.62\%) & 49,848 (T) + 16,859 (I) (28.95\%) \\
256 & 1 & 95.00 (T) + 100 (I) (62.66\%) & 56 (T) + 68 (I) (7.18\%) & 88,445 (T) + 26,001 (I) (24.84\%) & 61,485 (T) + 15,901 (I) (33.59\%) \\
256 & 2 & 95.00 (T) + 100 (I) (62.66\%) & 56 (T) + 71 (I) (7.35\%) & 88,767 (T) + 34,735 (I) (26.80\%) & 61,564 (T) + 20,656 (I) (35.69\%) \\
\hline
\multicolumn{6}{|c|}{\textbf{Hidden Nodes = 150}} \\
\hline
64  & 1 & 110.50 (T) + 45 (I) (49.84\%) & 59 (T) + 68 (I) (7.35\%) & 69,854 (T) + 13,244 (I) (18.03\%) & 57,558 (T) + 9,990 (I) (29.32\%) \\
64  & 2 & 110.50 (T) + 45 (I) (49.84\%) & 59 (T) + 71 (I) (7.52\%) & 70,107 (T) + 24,909 (I) (20.92\%) & 57,567 (T) + 16,717 (I) (32.24\%) \\
128 & 1 & 127.00 (T) + 79 (I) (66.03\%) & 59 (T) + 68 (I) (7.35\%) & 78,097 (T) + 17,828 (I) (20.82\%) & 62,410 (T) + 12,065 (I) (32.32\%) \\
128 & 2 & 127.00 (T) + 79 (I) (66.03\%) & 59 (T) + 71 (I) (7.52\%) & 78,308 (T) + 29,429 (I) (23.28\%) & 62,518 (T) + 18,732 (I) (35.24\%) \\
256 & 1 & 161.00 (T) + 145 (I) (98.08\%) & 59 (T) + 68 (I) (7.35\%) & 101,618 (T) + 26,020 (I) (27.70\%) & 75,616 (T) + 16,156 (I) (39.83\%) \\
256 & 2 & 161.00 (T) + 145 (I) (98.08\%) & 59 (T) + 71 (I) (7.52\%) & 101,899 (T) + 37,493 (I) (30.25\%) & 75,701 (T) + 22,381 (I) (42.57\%) \\
\hline
\end{tabular}
\vspace{1mm}
\begin{flushleft}
\small \textit{Note:} T = Training, I = Inference.
\end{flushleft}
\label{tab:training_resource}
\end{table*}

For the regression tasks on DCS, as shown in Fig.~\ref{fig:figure6_accuracy_dcs}(a)--(c), we take MATLAB built-in SVD as the reference; OJR-based SVD achieves comparable mean absolute error (MAE). The time consumption of OJR-SVD is overall higher as $N_0$, batch size, and $L$ increase than built-in SVD because the built-in SVD is deeply optimized by LAPACK matrix decomposition libraries~\cite{anderson1999lapack}, yet it is intractable to implement on a standalone FPGA SoC platform. Similar MAE trends can be observed in FLIM application (Fig.~\ref{fig:figure7_accuracy_flim}(a)--(c)). As we are interested in the scenario when batch size equals 1, Fig.~\ref{fig:figure6_accuracy_dcs}(e) and Fig.~\ref{fig:figure7_accuracy_flim}(e) indicate that the latency of naïve MATLAB implementation is more sensitive when batch size increases in OBT. Fig.~\ref{fig:figure6_accuracy_dcs}(d) and (j) indicate the relation between training latency and $N_0$; there is no significant latency increase beyond $L = 400$. In Fig.~\ref{fig:figure6_accuracy_dcs}(f) and Fig.~\ref{fig:figure7_accuracy_flim}(f), latency increases steadily after $L = 51$. But notably, as shown in Fig.~\ref{fig:figure6_accuracy_dcs}(g) and Fig.~\ref{fig:figure7_accuracy_flim}(c), the accuracy maintains high even at $L=50$. Table~\ref{tab:training_resource} shows the hardware utilization for training (T) and inference (I) of different OSOS-ELM topologies (\#IN, $L$, \#ON). The BRAM usage is proportional to \#IN, as more BRAMs are needed to store $\bm{W}$ and $\bm{b}$ between input and hidden layers. The number of used DSPs remains nearly consistent due to the same pipelining and unrolling strategy, with data dependencies limiting further parallelization. The numbers of flip-flops (FFs) and look-up tables (LUTs) are proportional to model size, as more memory is allocated to LUTs, and additional FFs are needed in pipelining. IT costs around 4 seconds on the PS, but this is not included in the latency evaluations, as only OBT is involved in online learning and is implemented on the PL. The latency of OBT and inference for each sample under different $L$, \#IN, and \#ON configurations is shown in Fig.~\ref{fig:figure8_GPU_compare}. The total number of training samples is 8{,}000, divided into 250 for IT and 7{,}750 for OBT, where the latter are consecutive input training vectors. Among the configuration parameters, $L$ is the most influential factor for latency, while increasing \#IN and \#ON results in minor changes.

\begin{table}[t]
\caption{Time consumption (s) for classifying 110{,}000 test histograms in foggy conditions using LiDAR}
\label{tab:lidar_runtime}
\centering
\scriptsize 
\renewcommand{\arraystretch}{1.1}
\setlength{\tabcolsep}{3.5pt} 
\begin{tabular}{lcccccc}
\toprule
\textbf{Method} & $L{=}100$ & $L{=}200$ & $L{=}300$ & $L{=}400$ & $L{=}500$ & $L{=}600$ \\
\midrule
Built-in SVD & 5.04 & 27.86 & 43.46 & 78.81 & 112.57 & 155.46 \\
OJR-SVD      & 28.06 & 77.99 & 131.27 & 261.87 & 508.68 & 881.65 \\
\bottomrule
\end{tabular}
\end{table}

In the GPU implementation, training latency is primarily determined by $L$, whereas \#IN and \#ON have negligible effects. Compared to the GPU implementation, FPGA-based OBT latency is more sensitive to changes in $L$. As $L$ increases, the latency of FPGA increases faster than that of GPU. When $L = 150$, the GPU shows significantly lower OBT latency compared to the FPGA, showing that GPUs are more efficient for large OSOS-ELM networks, while FPGAs are better suited for smaller networks. Across all tested configurations, inference on the GPU is consistently faster than on the FPGA. This is because GPUs excel in regular MVM for inference (e.g., Lines $\bm{32}$–$\bm{33}$), but are less efficient in performing iterative vector–scalar operations, as required in Lines $\bm{24}$–$\bm{30}$ for training. Although the overall power consumption of both FPGA and GPU is comparable, GPU inference consumes more power than GPU training when $L = 50$, which is opposite to FPGA behavior, where training consumes more energy than inference. This difference in power scaling becomes more pronounced as $L$ increases due to the complex and iterative computational nature of OBT training while inference comprises streamlined MVMs. Profiling using Nsight Systems reveals that OBT training induces frequent host-device synchronization, as kernel launch API calls are managed by the CPU. Across different workloads, the latency induced by kernel launches remains unchanged since the number of kernel launches is constant. However, for small $L$ (e.g., 50), GPU computation completes so quickly that it often has to stall for the CPU to launch the next kernel. As $L$ grows (e.g., 150), the computational workload dominates, and the API calls by CPU are hidden. As a result, GPU active time and utilization are increased, which in turn increases power consumption during training.

In practical scenarios, our post-processing OSOS-ELM hardware is interfaced with front-end DCS and FLIM systems. We validated its capability to process front-end sensing data using the latest on-chip $g_2(\tau)$~\cite{della2023field} for DCS and $h(t)$ for FLIM~\cite{xiao2021dynamic}. Here, hardware interfacing latency is not considered. For performance evaluation, we take the latest on-chip DCS autocorrelator~\cite{della2023field} as an example. This autocorrelator generates an ensembled $g_2(\tau)$ with 16 linearly spaced lags within a 60~ms integration time at the highest SPAD array frame rate. A total of 100 repeated measurements (lasting 6~s) are performed to obtain a high-SNR ensemble of $g_2(\tau)$. We configure the OSOS-ELM IP core with \#IN~=~16, $L = 50$, and \#ON~=~2, as shown in Fig.~\ref{fig:figure6_accuracy_dcs}(c), where no significant accuracy degradation is observed at $L = 50$. The training and inference latencies of the IP core, measured using an on-chip timer, are 0.11~ms and 0.018~ms, respectively—both of which are significantly shorter than the integration time. Therefore, without considering the latency of hardware interfacing between the autocorrelator and the OSOS-ELM, the IP core is fully capable of processing ensembled $g_2(\tau)$. However, the training and inference speed of our hardware is not fast enough to keep up with the latest on-chip histogramming FLIM system~\cite{erdogan2019cmos}, which achieves a high histogram throughput of 14.2~k lines per second. Thus, data caching using DDR or off-chip memory is essential to handle such throughput effectively.

\begin{table*}[t]
    \caption{Comparison of existing FPGA-based ELM algorithms and hardware implementations}
    \label{tab:comparison_elm_fpga}
    \centering
    \renewcommand{\arraystretch}{1.2}
    \resizebox{\textwidth}{!}{%
    \begin{tabular}{lccccc}
        \toprule
        \textbf{} & 
        \begin{tabular}{@{}c@{}}\textbf{OS-ELM-FPGA-}\\\textit{Electronics~\cite{frances2018moving}}\end{tabular} &
        \begin{tabular}{@{}c@{}}\textbf{ELM-FPGA-}\\\textit{TII~\cite{bataller2016support}}\end{tabular} &
        \begin{tabular}{@{}c@{}}\textbf{OS-ELM-FPGA-}\\\textit{TCAD~\cite{safaei2018system}}\end{tabular} &
        \begin{tabular}{@{}c@{}}\textbf{OS-ELM-FPGA-}\\\textit{TC~\cite{tsukada2020neural}}\end{tabular} &
        \textbf{OSOS-ELM (This work)} \\
        \midrule
        Platform & Xilinx Virtex-7 & Xilinx Spartan6 \& Virtex-7 & Xilinx ZYNQ-7000 & Xilinx PYNQ-Z2 & \textbf{Xilinx ZCU104 MPSoC} \\
        Method solving matrix inversion & N.A.\textsuperscript{a} & QRD & QRD & Forgetting Mechanism & \textbf{OJR-SVD} \\
        Technology & 28 nm & 28 nm \& 45 nm & 28 nm & 28 nm & \textbf{16 nm} \\
        Application & \begin{tabular}{@{}c@{}}Image Seg.\\(classification)\end{tabular} &
        \begin{tabular}{@{}c@{}}Brain Area Detection\\(Classification)\end{tabular} &
        \begin{tabular}{@{}c@{}}Human Action Recognition\\(Classification)\end{tabular} &
        \begin{tabular}{@{}c@{}}Anomaly Detector\\(Classification)\end{tabular} &
        \begin{tabular}{@{}c@{}}\textbf{LiDAR, FLIM, DCS}\\\textbf{(Cls. + Regr.)}\end{tabular} \\
        Implementation language & HLS & RTL & RTL & HLS & \textbf{HLS} \\
        Data format & FXP & FXP & FXP & FXP & \textbf{FLP 64-b} \\
        Support IT & No & N.A. & Yes & No & \textbf{Yes} \\
        Transferable to other applications & N.A. & N.A. & N.A. & N.A. & \textbf{Yes} \\
        Topology (\#IN, $L$, \#ON) & (100, 150, 7) & (17, 28, 3) & (40, 40, 13) & (128, 64, 1) & \textbf{(128, 150, 2)} \\
        \# Parameters & 16,200 & 588 & 2,160 & 8,320 & \textbf{19,650} \\
       Task & Training \& Inference & Training \& Inference & Training \& Inference & Training \& Inference & \textbf{Training \& Inference} \\
        Clock Frequency & Overall 250 MHz (24b FXP) &
        \begin{tabular}{@{}c@{}} Spartan-6: Overall 30 MHz (30b FXP)\\Virtex-7: Overall 65 MHz (30b FXP)\end{tabular} &
        \text{Overall 65 MHz (30b FXP)} & \text{Overall 100 MHz (32b FXP)} &
        \textbf{101.2 \& 61.53 MHz (FLP)} \\
        Power & N.A. & N.A. & 1.423 W (total) & 3.1 W (total) & \textbf{4.615 \& 4.193 W} \\
        Latency (per sample) & 0.56 ms \& N.A. & \begin{tabular}{@{}c@{}}Spartan-6: 11.7 ms \& 0.03 ms\\Virtex-7: 5.3 ms \& 0.014 ms \end{tabular} & 0.12 ms \& 0.02 ms & 0.22 ms \& 0.10 ms & \textbf{1.05 ms \& 0.18 ms} \\
        \bottomrule
    \end{tabular}
    } 
    \vspace{1mm}
    \begin{flushleft}
    \footnotesize%
    \textsuperscript{a}This work does not support IT; no matrix pseudo-inversion.\\
    \textsuperscript{b}All topologies listed are reconfigurable and evaluated in the corresponding papers.
    \end{flushleft}
\end{table*}

\textbf{Table~\ref{tab:comparison_elm_fpga}} compares the existing hardware-oriented ELM algorithms and FPGA implementations. Distinguished from prior work, we conduct two FLP-based regression tasks that are more sensitive to noise. We demonstrate that different \#fraction bit settings affect the accuracy across applications differently, making FXP less transferable to other use cases. Our approach achieves a trade-off between latency and generalization. Notably, the evaluated model has the largest size with the most parameters. Despite this, real-time per-sample training and inference can be achieved.

We attempted to implement the model on a PYNQ-Z2 FPGA. However, the hardware resources were insufficient to complete the place-and-route process for a large OSOS-ELM with high \#IN and $L$. As a result, OSOS-ELM is implemented on a more advanced FPGA platform, offering the highest resource capacity among existing work. Another justification for using an advanced FPGA is that our task involves regression, which is highly sensitive to FXP quantization errors. A further limitation of FXP for OSOS-ELM—or ELM in general—is that the trained weight matrix $\bm{\eta}$ can include very large values, potentially in the hundreds or thousands. FXP with a limited number of integer bits cannot accurately represent such large values and must round them to the maximum value representable. This introduces rounding errors that affect precision. FLP, by contrast, ensures high numerical precision, though it results in increased hardware utilization and longer critical paths. In classification tasks, however, the decision boundaries are discrete, allowing small rounding deviations to be tolerated. Therefore, using an FPGA with more resources enables the deployment of larger, more versatile models. Naturally, more power is consumed—approximately 73\% of the total power consumption is attributed to the PS side.

\section{conclusion}
This work is the first to demonstrate the arithmetic efficiency and FPGA-based hardware prototype of the proposed OSOS-ELM algorithm through three case studies in encoded single-photon optics: regression tasks for fluorescence lifetime reconstruction in FLIM and BFi reconstruction in DCS, as well as a classification task in LiDAR. A hardware-efficient OJR-SVD is integrated into the OSOS-ELM to efficiently handle computationally intensive matrix inversion, leveraging task-level parallelism. Training and inference datasets are generated using realistic analytical models for performance evaluation. The OSOS-ELM algorithm is compared with built-in SVD-based matrix inversion approaches, showing comparable accuracy and latency in software while highlighting its potential for high parallelism in FPGA implementation.

For hardware implementation, we conducted extensive evaluations of model topologies, data formats, on-hardware computing latency, power consumption, and hardware utilization. Additionally, we propose a holistic and generalized computing platform for OSOS-ELM with detailed memory segmentation and workload scheduling. This platform enables real-time training and inference per sample, bridging the gap to practical experiments. A GPU-based OSOS-ELM solution is also proposed to explore alternative parallelism strategies using CUDA. Results suggest that FPGA and GPU are suitable for compact and large OSOS-ELM modules, respectively. Both FPGA and GPU implementations are highly transferable to other applications, as the hardware implementation is highly parameterized. While this study focuses on regression tasks, the algorithm and hardware architecture can be adapted for other time-series classification tasks by adjusting the loss function and sorting labels.

\section*{Acknowledgment}
This work is supported by the EPSRC (EP/T00097X/1); the Quantum Technology Hub in Quantum Imaging (QuantiC), and the University of Strathclyde. Xingda Li also acknowledges support from China Scholarship Council.

\bibliographystyle{IEEEtran}
\bibliography{bibtex/bib/IEEEabrv.bib,bibtex/bib/IEEEexample.bib}{}

\begin{thebibliography}{10}
\providecommand{\url}[1]{#1}
\csname url@samestyle\endcsname
\providecommand{\newblock}{\relax}
\providecommand{\bibinfo}[2]{#2}
\providecommand{\BIBentrySTDinterwordspacing}{\spaceskip=0pt\relax}
\providecommand{\BIBentryALTinterwordstretchfactor}{4}
\providecommand{\BIBentryALTinterwordspacing}{\spaceskip=\fontdimen2\font plus
\BIBentryALTinterwordstretchfactor\fontdimen3\font minus \fontdimen4\font\relax}
\providecommand{\BIBforeignlanguage}[2]{{%
\expandafter\ifx\csname l@#1\endcsname\relax
\typeout{** WARNING: IEEEtran.bst: No hyphenation pattern has been}%
\typeout{** loaded for the language `#1'. Using the pattern for}%
\typeout{** the default language instead.}%
\else
\language=\csname l@#1\endcsname
\fi
#2}}
\providecommand{\BIBdecl}{\relax}
\BIBdecl

\bibitem{Sparks2021}
\BIBentryALTinterwordspacing
P.~Sparks, ``The route to a trillion devices,'' 2021, accessed: 2025-04-07. [Online]. Available: \url{https://community.arm.com/iot/b/blog/posts/whitepaper-the-route-to-a-trillion-devices}
\BIBentrySTDinterwordspacing

\bibitem{lai2018cmsis}
L.~Lai, N.~Suda, and V.~Chandra, ``Cmsis-nn: Efficient neural network kernels for arm cortex-m cpus,'' \emph{arXiv preprint arXiv:1801.06601}, 2018.

\bibitem{lin2022device}
J.~Lin, L.~Zhu, W.-M. Chen, W.-C. Wang, C.~Gan, and S.~Han, ``On-device training under 256kb memory,'' \emph{Advances in Neural Information Processing Systems}, vol.~35, pp. 22\,941--22\,954, 2022.

\bibitem{khare2022vhers}
S.~K. Khare, N.~B. Gaikwad, and V.~Bajaj, ``Vhers: a novel variational mode decomposition and hilbert transform-based eeg rhythm separation for automatic adhd detection,'' \emph{IEEE Transactions on Instrumentation and Measurement}, vol.~71, pp. 1--10, 2022.

\bibitem{chen2023s}
Y.-T. Chen, Y.-C. Chuang, L.-S. Chang, and A.-Y. Wu, ``S-qrd-elm: scalable qr-decomposition-based extreme learning machine engine supporting online class-incremental learning for ecg-based user identification,'' \emph{IEEE Transactions on Circuits and Systems I: Regular Papers}, vol.~70, no.~6, pp. 2342--2355, 2023.

\bibitem{mahuli2016prediction}
A.~P. Mahuli, B.~N. Krupa, and K.~George, ``Prediction of photoplethysmogram signals using sequential elm,'' in \emph{2016 International Joint Conference on Neural Networks (IJCNN)}.\hskip 1em plus 0.5em minus 0.4em\relax IEEE, 2016, pp. 1977--1984.

\bibitem{shi2021novel}
X.~Shi, Q.~Kang, J.~An, and M.~Zhou, ``Novel l1 regularized extreme learning machine for soft-sensing of an industrial process,'' \emph{IEEE Transactions on Industrial Informatics}, vol.~18, no.~2, pp. 1009--1017, 2021.

\bibitem{zhao2018novel}
C.~Zhao, K.~Li, Y.~Li, L.~Wang, Y.~Luo, X.~Xu, X.~Ding, and Q.~Meng, ``Novel method based on variational mode decomposition and a random discriminative projection extreme learning machine for multiple power quality disturbance recognition,'' \emph{IEEE Transactions on Industrial Informatics}, vol.~15, no.~5, pp. 2915--2926, 2018.

\bibitem{zang2024object}
Z.~Zang and D.~D. Uei~Li, ``Object classification through heterogeneous fog with a fast data-driven algorithm using a low-cost single-photon avalanche diode array,'' \emph{Optics Express}, vol.~32, no.~19, pp. 33\,294--33\,304, 2024.

\bibitem{zang2022fast}
Z.~Zang, D.~Xiao, Q.~Wang, Z.~Li, W.~Xie, Y.~Chen, and D.~D.~U. Li, ``Fast analysis of time-domain fluorescence lifetime imaging via extreme learning machine,'' \emph{Sensors}, vol.~22, no.~10, p. 3758, 2022.

\bibitem{zang2025fast}
Z.~Zang, M.~Pan, Y.~Zhang, and D.~D.~U. Li, ``Fast blood flow index reconstruction of diffuse correlation spectroscopy using a back-propagation-free data-driven algorithm,'' \emph{Biomedical Optics Express}, vol.~16, no.~3, pp. 1254--1269, 2025.

\bibitem{tang2015extreme}
J.~Tang, C.~Deng, and G.-B. Huang, ``Extreme learning machine for multilayer perceptron,'' \emph{IEEE transactions on neural networks and learning systems}, vol.~27, no.~4, pp. 809--821, 2015.

\bibitem{huang2005line}
G.-B. Huang, N.-Y. Liang, H.-J. Rong, P.~Saratchandran, and N.~Sundararajan, ``On-line sequential extreme learning machine.'' \emph{Computational Intelligence}, vol. 2005, pp. 232--237, 2005.

\bibitem{torrey2010transfer}
L.~Torrey and J.~Shavlik, ``Transfer learning,'' in \emph{Handbook of research on machine learning applications and trends: algorithms, methods, and techniques}.\hskip 1em plus 0.5em minus 0.4em\relax IGI global, 2010, pp. 242--264.

\bibitem{takata2021implementation}
M.~Takata, S.~Araki, K.~Kimura, and Y.~Nakamura, ``On an implementation of the one-sided jacobi method with high accuracy,'' in \emph{Advances in Parallel \& Distributed Processing, and Applications: Proceedings from PDPTA'20, CSC'20, MSV'20, and GCC'20}.\hskip 1em plus 0.5em minus 0.4em\relax Springer, 2021, pp. 713--724.

\bibitem{becker2012fluorescence}
W.~Becker, ``Fluorescence lifetime imaging--techniques and applications,'' \emph{Journal of microscopy}, vol. 247, no.~2, pp. 119--136, 2012.

\bibitem{durduran2014diffuse}
T.~Durduran and A.~G. Yodh, ``Diffuse correlation spectroscopy for non-invasive, micro-vascular cerebral blood flow measurement,'' \emph{Neuroimage}, vol.~85, pp. 51--63, 2014.

\bibitem{chen2020data}
G.~Chen, C.~Wiede, and R.~Kokozinski, ``Data processing approaches on spad-based d-tof lidar systems: A review,'' \emph{IEEE Sensors Journal}, vol.~21, no.~5, pp. 5656--5667, 2020.

\bibitem{xiao2021dynamic}
D.~Xiao, Z.~Zang, N.~Sapermsap, Q.~Wang, W.~Xie, Y.~Chen, and D.~D. Uei~Li, ``Dynamic fluorescence lifetime sensing with cmos single-photon avalanche diode arrays and deep learning processors,'' \emph{Biomedical Optics Express}, vol.~12, no.~6, pp. 3450--3462, 2021.

\bibitem{della2023field}
F.~M. Della~Rocca, E.~J. Sie, R.~Catoen, F.~Marsili, and R.~K. Henderson, ``Field programmable gate array compression for large array multispeckle diffuse correlation spectroscopy,'' \emph{Journal of Biomedical Optics}, vol.~28, no.~5, pp. 057\,001--057\,001, 2023.

\bibitem{della2020128}
F.~M. Della~Rocca, H.~Mai, S.~W. Hutchings, T.~Al~Abbas, K.~Buckbee, A.~Tsiamis, P.~Lomax, I.~Gyongy, N.~A. Dutton, and R.~K. Henderson, ``A 128$\times$ 128 spad motion-triggered time-of-flight image sensor with in-pixel histogram and column-parallel vision processor,'' \emph{IEEE Journal of Solid-State Circuits}, vol.~55, no.~7, pp. 1762--1775, 2020.

\bibitem{henderson2019192}
R.~K. Henderson, N.~Johnston, F.~M. Della~Rocca, H.~Chen, D.~D.-U. Li, G.~Hungerford, R.~Hirsch, D.~Mcloskey, P.~Yip, and D.~J. Birch, ``A $192$ $\times128 $ time correlated spad image sensor in 40-nm cmos technology,'' \emph{IEEE Journal of Solid-State Circuits}, vol.~54, no.~7, pp. 1907--1916, 2019.

\bibitem{gyongy2020high}
I.~Gyongy, S.~W. Hutchings, A.~Halimi, M.~Tyler, S.~Chan, F.~Zhu, S.~McLaughlin, R.~K. Henderson, and J.~Leach, ``High-speed 3d sensing via hybrid-mode imaging and guided upsampling,'' \emph{Optica}, vol.~7, no.~10, pp. 1253--1260, 2020.

\bibitem{portaluppi2022multi}
D.~Portaluppi, K.~Pasquinelli, I.~Cusini, and F.~Zappa, ``Multi-channel fpga time-to-digital converter with 10 ps bin and 40 ps fwhm,'' \emph{IEEE Transactions on Instrumentation and Measurement}, vol.~71, pp. 1--9, 2022.

\bibitem{li2009fpga}
D.-U. Li, R.~Walker, J.~Richardson, B.~Rae, A.~Buts, D.~Renshaw, and R.~Henderson, ``Fpga implementation of a video-rate fluorescence lifetime imaging system with a 32$\times$ 32 cmos single-photon avalanche diode array,'' in \emph{2009 IEEE International Symposium on Circuits and Systems}.\hskip 1em plus 0.5em minus 0.4em\relax IEEE, 2009, pp. 3082--3085.

\bibitem{wayne2023massively}
M.~A. Wayne, E.~J. Sie, A.~C. Ulku, P.~Mos, A.~Ardelean, F.~Marsili, C.~Bruschini, and E.~Charbon, ``Massively parallel, real-time multispeckle diffuse correlation spectroscopy using a 500$\times$ 500 spad camera,'' \emph{Biomedical Optics Express}, vol.~14, no.~2, pp. 703--713, 2023.

\bibitem{zang2023compact}
Z.~Zang, D.~Xiao, Q.~Wang, Z.~Jiao, Y.~Chen, and D.~D.~U. Li, ``Compact and robust deep learning architecture for fluorescence lifetime imaging and fpga implementation,'' \emph{Methods and Applications in Fluorescence}, vol.~11, no.~2, p. 025002, 2023.

\bibitem{smith2019fast}
J.~T. Smith, R.~Yao, N.~Sinsuebphon, A.~Rudkouskaya, N.~Un, J.~Mazurkiewicz, M.~Barroso, P.~Yan, and X.~Intes, ``Fast fit-free analysis of fluorescence lifetime imaging via deep learning,'' \emph{Proceedings of the national academy of sciences}, vol. 116, no.~48, pp. 24\,019--24\,030, 2019.

\bibitem{chen2022generative}
Y.-I. Chen, Y.-J. Chang, S.-C. Liao, T.~D. Nguyen, J.~Yang, Y.-A. Kuo, S.~Hong, Y.-L. Liu, H.~G. Rylander~III, S.~R. Santacruz \emph{et~al.}, ``Generative adversarial network enables rapid and robust fluorescence lifetime image analysis in live cells,'' \emph{Communications Biology}, vol.~5, no.~1, p.~18, 2022.

\bibitem{chen2023spatial}
Y.-I. Chen, Y.-J. Chang, Y.~Sun, S.-C. Liao, S.~R. Santacruz, and H.-C. Yeh, ``Spatial resolution enhancement in photon-starved sted imaging using deep learning-based fluorescence lifetime analysis,'' \emph{Nanoscale}, vol.~15, no.~21, pp. 9449--9456, 2023.

\bibitem{wang2024quantification}
Q.~Wang, M.~Pan, Z.~Zang, and D.~D.-U. Li, ``Quantification of blood flow index in diffuse correlation spectroscopy using a robust deep learning method,'' \emph{Journal of Biomedical Optics}, vol.~29, no.~1, pp. 015\,004--015\,004, 2024.

\bibitem{afshar2020event}
S.~Afshar, T.~J. Hamilton, L.~Davis, A.~Van~Schaik, and D.~Delic, ``Event-based processing of single photon avalanche diode sensors,'' \emph{IEEE Sensors Journal}, vol.~20, no.~14, pp. 7677--7691, 2020.

\bibitem{lin2024coupling}
Y.~Lin, P.~Mos, A.~Ardelean, C.~Bruschini, and E.~Charbon, ``Coupling a recurrent neural network to spad tcspc systems for real-time fluorescence lifetime imaging,'' \emph{Scientific Reports}, vol.~14, no.~1, p. 3286, 2024.

\bibitem{mai2023development}
H.~Mai, A.~Jarman, A.~T. Erdogan, C.~Treacy, N.~Finlayson, R.~K. Henderson, and S.~P. Poland, ``Development of a high-speed line-scanning fluorescence lifetime imaging microscope for biological imaging,'' \emph{Optics letters}, vol.~48, no.~8, pp. 2042--2045, 2023.

\bibitem{frances2018moving}
J.~V. Frances-Villora, A.~Rosado-Mu{\~n}oz, M.~Bataller-Mompean, J.~Barrios-Aviles, and J.~F. Guerrero-Martinez, ``Moving learning machine towards fast real-time applications: A high-speed fpga-based implementation of the os-elm training algorithm,'' \emph{Electronics}, vol.~7, no.~11, p. 308, 2018.

\bibitem{safaei2018system}
A.~Safaei, Q.~J. Wu, T.~Akilan, and Y.~Yang, ``System-on-a-chip (soc)-based hardware acceleration for an online sequential extreme learning machine (os-elm),'' \emph{IEEE transactions on computer-aided design of integrated circuits and systems}, vol.~38, no.~11, pp. 2127--2138, 2018.

\bibitem{tsukada2020neural}
M.~Tsukada, M.~Kondo, and H.~Matsutani, ``A neural network-based on-device learning anomaly detector for edge devices,'' \emph{IEEE Transactions on Computers}, vol.~69, no.~7, pp. 1027--1044, 2020.

\bibitem{he2021low}
Z.~He, C.~Shi, T.~Wang, Y.~Wang, M.~Tian, X.~Zhou, P.~Li, L.~Liu, N.~Wu, and G.~Luo, ``A low-cost fpga implementation of spiking extreme learning machine with on-chip reward-modulated stdp learning,'' \emph{IEEE Transactions on Circuits and Systems II: Express Briefs}, vol.~69, no.~3, pp. 1657--1661, 2021.

\bibitem{decherchi2012efficient}
S.~Decherchi, P.~Gastaldo, A.~Leoncini, and R.~Zunino, ``Efficient digital implementation of extreme learning machines for classification,'' \emph{IEEE Transactions on Circuits and Systems II: Express Briefs}, vol.~59, no.~8, pp. 496--500, 2012.

\bibitem{sahani2019fpga}
M.~Sahani and P.~K. Dash, ``Fpga-based online power quality disturbances monitoring using reduced-sample hht and class-specific weighted rvfln,'' \emph{IEEE Transactions on Industrial Informatics}, vol.~15, no.~8, pp. 4614--4623, 2019.

\bibitem{sahani2019automatic}
------, ``Automatic power quality events recognition using modes decomposition based online p-norm adaptive extreme learning machine,'' \emph{IEEE Transactions on Industrial Informatics}, vol.~16, no.~7, pp. 4355--4364, 2019.

\bibitem{huang2021generic}
H.~Huang, J.~Yang, H.-J. Rong, and S.~Du, ``A generic fpga-based hardware architecture for recursive least mean p-power extreme learning machine,'' \emph{Neurocomputing}, vol. 456, pp. 421--435, 2021.

\bibitem{chuang2021arbitrarily}
Y.-C. Chuang, Y.-T. Chen, H.-T. Li, and A.-Y.~A. Wu, ``An arbitrarily reconfigurable extreme learning machine inference engine for robust ecg anomaly detection,'' \emph{IEEE Open Journal of Circuits and Systems}, vol.~2, pp. 196--209, 2021.

\bibitem{chen2015128}
Y.~Chen, E.~Yao, and A.~Basu, ``A 128-channel extreme learning machine-based neural decoder for brain machine interfaces,'' \emph{IEEE transactions on biomedical circuits and systems}, vol.~10, no.~3, pp. 679--692, 2015.

\bibitem{bataller2016support}
M.~Bataller-Mompe{\'a}n, J.~M. Mart{\'\i}nez-Villena, A.~Rosado-Munoz, J.~V. Franc{\'e}s-V{\'\i}llora, J.~F. Guerrero-Mart{\'\i}nez, M.~Wegrzyn, and M.~Adamski, ``Support tool for the combined software/hardware design of on-chip elm training for slff neural networks,'' \emph{IEEE Transactions on Industrial Informatics}, vol.~12, no.~3, pp. 1114--1123, 2016.

\bibitem{li2019robust}
H.-T. Li, C.-Y. Chou, Y.-T. Chen, S.-H. Wang, and A.-Y. Wu, ``Robust and lightweight ensemble extreme learning machine engine based on eigenspace domain for compressed learning,'' \emph{IEEE Transactions on Circuits and Systems I: Regular Papers}, vol.~66, no.~12, pp. 4699--4712, 2019.

\bibitem{liang2019memristive}
H.~Liang, H.~Cheng, J.~Wei, L.~Zhang, L.~Yang, Y.~Zhao, and H.~Guo, ``Memristive neural networks: A neuromorphic paradigm for extreme learning machine,'' \emph{IEEE Transactions on Emerging Topics in Computational Intelligence}, vol.~3, no.~1, pp. 15--23, 2019.

\bibitem{dong2021neuromorphic}
Z.~Dong, C.~S. Lai, Z.~Zhang, D.~Qi, M.~Gao, and S.~Duan, ``Neuromorphic extreme learning machines with bimodal memristive synapses,'' \emph{Neurocomputing}, vol. 453, pp. 38--49, 2021.

\bibitem{li2010real}
D.-U. Li, J.~Arlt, J.~Richardson, R.~Walker, A.~Buts, D.~Stoppa, E.~Charbon, and R.~Henderson, ``Real-time fluorescence lifetime imaging system with a 32$\times$ 32 0.13 $\mu$m cmos low dark-count single-photon avalanche diode array,'' \emph{Optics express}, vol.~18, no.~10, pp. 10\,257--10\,269, 2010.

\bibitem{mai2020flow}
H.~Mai, S.~P. Poland, F.~M. Della~Rocca, C.~Treacy, J.~Aluko, J.~Nedbal, A.~T. Erdogan, I.~Gyongy, R.~Walker, S.~M. Ameer-Beg \emph{et~al.}, ``Flow cytometry visualization and real-time processing with a cmos spad array and high-speed hardware implementation algorithm,'' in \emph{Imaging, Manipulation, and Analysis of Biomolecules, Cells, and Tissues XVIII}, vol. 11243.\hskip 1em plus 0.5em minus 0.4em\relax SPIE, 2020, pp. 31--37.

\bibitem{lin2019diffuse}
W.~Lin, D.~R. Busch, C.~C. Goh, J.~Barsi, and T.~F. Floyd, ``Diffuse correlation spectroscopy analysis implemented on a field programmable gate array,'' \emph{IEEE Access}, vol.~7, pp. 122\,503--122\,512, 2019.

\bibitem{buchholz2012fpga}
J.~Buchholz, J.~W. Krieger, G.~Mocs{\'a}r, B.~Kreith, E.~Charbon, G.~V{\'a}mosi, U.~Kebschull, and J.~Langowski, ``Fpga implementation of a 32x32 autocorrelator array for analysis of fast image series,'' \emph{Optics express}, vol.~20, no.~16, pp. 17\,767--17\,782, 2012.

\bibitem{zang2025towards}
Z.~Zang, Q.~Wang, M.~Pan, Y.~Zhang, X.~Chen, X.~Li, and D.~D.~U. Li, ``Towards high-performance deep learning architecture and hardware accelerator design for robust analysis in diffuse correlation spectroscopy,'' \emph{Computer Methods and Programs in Biomedicine}, vol. 258, p. 108471, 2025.

\bibitem{sun2022multi}
M.~Sun, S.~Zhuo, and P.~Y. Chiang, ``Multi-scale histogram-based probabilistic deep neural network for super-resolution 3d lidar imaging,'' \emph{Sensors}, vol.~23, no.~1, p. 420, 2022.

\bibitem{cui202380}
L.~Cui, J.~Li, S.~Zhuo, Y.~Wu, S.~Zhou, J.~Qian, M.~Sun, J.~Wang, P.~Y. Chiang, and Y.~Chen, ``80$\times$ 120 ai-enhanced lidar system based on a lightweight intensity--rgb--dtof sensor fusion neural network deployed on an edge device,'' \emph{Optics Letters}, vol.~48, no.~23, pp. 6192--6195, 2023.

\bibitem{huang2006extreme}
G.-B. Huang, Q.-Y. Zhu, and C.-K. Siew, ``Extreme learning machine: theory and applications,'' \emph{Neurocomputing}, vol.~70, no. 1-3, pp. 489--501, 2006.

\bibitem{alessandrini2020singular}
M.~Alessandrini, G.~Biagetti, P.~Crippa, L.~Falaschetti, L.~Manoni, and C.~Turchetti, ``Singular value decomposition in embedded systems based on arm cortex-m architecture,'' \emph{Electronics}, vol.~10, no.~1, p.~34, 2020.

\bibitem{demmel1992jacobi}
J.~Demmel and K.~Veseli{\'c}, ``Jacobi’s method is more accurate than qr,'' \emph{SIAM journal on matrix analysis and applications}, vol.~13, no.~4, pp. 1204--1245, 1992.

\bibitem{poon2020deep}
C.-S. Poon, F.~Long, and U.~Sunar, ``Deep learning model for ultrafast quantification of blood flow in diffuse correlation spectroscopy,'' \emph{Biomedical Optics Express}, vol.~11, no.~10, pp. 5557--5564, 2020.

\bibitem{fereidouni2017rapid}
F.~Fereidouni, D.~Gorpas, D.~Ma, H.~Fatakdawala, and L.~Marcu, ``Rapid fluorescence lifetime estimation with modified phasor approach and laguerre deconvolution: a comparative study,'' \emph{Methods and applications in fluorescence}, vol.~5, no.~3, p. 035003, 2017.

\bibitem{li2020investigations}
Y.~Li, S.~Natakorn, Y.~Chen, M.~Safar, M.~Cunningham, J.~Tian, and D.~D.-U. Li, ``Investigations on average fluorescence lifetimes for visualizing multi-exponential decays,'' \emph{Frontiers in physics}, vol.~8, p. 576862, 2020.

\bibitem{boas1997spatially}
D.~A. Boas and A.~G. Yodh, ``Spatially varying dynamical properties of turbid media probed with diffusing temporal light correlation,'' \emph{Journal of the Optical Society of America A}, vol.~14, no.~1, pp. 192--215, 1997.

\bibitem{rice1944mathematical}
S.~O. Rice, ``Mathematical analysis of random noise,'' \emph{The Bell System Technical Journal}, vol.~23, no.~3, pp. 282--332, 1944.

\bibitem{helton2022numerical}
M.~Helton, S.~Rajasekhar, S.~Zerafa, K.~Vishwanath, and M.-A. Mycek, ``Numerical approach to quantify depth-dependent blood flow changes in real-time using the diffusion equation with continuous-wave and time-domain diffuse correlation spectroscopy,'' \emph{Biomedical Optics Express}, vol.~14, no.~1, pp. 367--384, 2022.

\bibitem{dong2012simultaneously}
L.~Dong, L.~He, Y.~Lin, Y.~Shang, and G.~Yu, ``Simultaneously extracting multiple parameters via fitting one single autocorrelation function curve in diffuse correlation spectroscopy,'' \emph{IEEE Transactions on Biomedical Engineering}, vol.~60, no.~2, pp. 361--368, 2012.

\bibitem{Nickolls2008Scalable}
\BIBentryALTinterwordspacing
J.~Nickolls, I.~Buck, M.~Garland, and K.~Skadron, ``Scalable parallel programming with cuda,'' in \emph{ACM SIGGRAPH 2008 Classes}, ser. SIGGRAPH '08.\hskip 1em plus 0.5em minus 0.4em\relax New York, NY, USA: Association for Computing Machinery, 2008. [Online]. Available: \url{https://doi.org/10.1145/1401132.1401152}
\BIBentrySTDinterwordspacing

\bibitem{holly2020profiling}
S.~Holly, A.~Wendt, and M.~Lechner, ``Profiling energy consumption of deep neural networks on nvidia jetson nano,'' in \emph{2020 11th International Green and Sustainable Computing Workshops (IGSC)}.\hskip 1em plus 0.5em minus 0.4em\relax IEEE, 2020, pp. 1--6.

\bibitem{anderson1999lapack}
E.~Anderson, Z.~Bai, C.~Bischof, L.~S. Blackford, J.~Demmel, J.~Dongarra, J.~Du~Croz, A.~Greenbaum, S.~Hammarling, A.~McKenney \emph{et~al.}, \emph{LAPACK users' guide}.\hskip 1em plus 0.5em minus 0.4em\relax SIAM, 1999.

\bibitem{erdogan2019cmos}
A.~T. Erdogan, R.~Walker, N.~Finlayson, N.~Krstaji{\'c}, G.~Williams, J.~Girkin, and R.~Henderson, ``A cmos spad line sensor with per-pixel histogramming tdc for time-resolved multispectral imaging,'' \emph{IEEE Journal of Solid-State Circuits}, vol.~54, no.~6, pp. 1705--1719, 2019.

\end{thebibliography}

\end{document}